\PassOptionsToPackage{table,dvipsnames}{xcolor}
\documentclass[]{amap}

\usepackage[toc,page,header]{appendix}
\usepackage{solarized-light}
\usepackage{multirow}
\usepackage{graphicx}
\usepackage{siunitx,array}
\usepackage{makecell,array,booktabs}
\usepackage{framed}

\usepackage{amsmath}
\usepackage{amsfonts}
\usepackage{placeins}
\usepackage[colorinlistoftodos]{todonotes}
\usepackage{longtable}
\usepackage{hhline}
\usepackage{fancyvrb}
\usepackage{float}
\usepackage{fvextra}
\usepackage{CJKutf8}
\usepackage{multicol}
\usepackage{tablefootnote}
\usepackage{threeparttable}
\usepackage{mdframed}
\usepackage{subcaption}
\usepackage[usestackEOL]{stackengine}
\usepackage[numbers]{natbib}
\newcommand{\commentout}[1]{}
\usepackage{enumitem}
\setlist[itemize]{leftmargin=15pt}
\usepackage{tabularx}
\usepackage{adjustbox}
\usepackage{colortbl} 
\usepackage{arydshln}
\usepackage{pifont}
\usepackage{bbding}
\definecolor{ampblue}{rgb}{0.82, 0.88, 0.94}
\usepackage{amssymb} 

\definecolor{groupgray}{gray}{0.94}
\definecolor{pdmsgray}{gray}{0.92}
\colorlet{oursblue}{ampblue}

\newcommand{\trajectorylegend}{The \textcolor{red}{\textbf{red}} curve marks the trajectory generated by PerceptDrive and the \textcolor{green!55!black}{\textbf{green}} curve the logged human trajectory, both projected onto the front and bev view.}
\usepackage{wrapfig}

\usepackage{hyperref}
\usepackage{cleveref}
\usepackage{tikz}
\usetikzlibrary{arrows.meta,positioning,fit,calc}

\hypersetup{
  colorlinks=true,
  linkcolor=blue!55!black,
  citecolor=blue!55!black,
  urlcolor=blue!55!black
}

\RequirePackage{xspace}
\makeatletter
\DeclareRobustCommand\onedot{\futurelet\@let@token\@onedot}
\def\@onedot{\ifx\@let@token.\else.\null\fi\xspace}

\makeatother

\makeatletter
\let\ftype@table\ftype@figure
\makeatother

\definecolor{abot1}{HTML}{0185FE}
\definecolor{abot2}{HTML}{0185FE}
\definecolor{abot3}{HTML}{0185FE}
\definecolor{abot4}{HTML}{0185FE}
\definecolor{abot5}{HTML}{FB8C00}
\definecolor{abot6}{HTML}{FB8C00}
\definecolor{abot7}{HTML}{FB8C00}

\title{PerceptDrive: Perception Prior World-Action Modeling with Adaptive Expert Routing for End-to-End Autonomous Driving}

\newcommand{\internshipwork}{\textsuperscript{\rm ,*}}
\newcommand{\projectleader}{\textsuperscript{\rm ,\ddag}}
\newcommand{\correspondingauthor}{\textsuperscript{\rm ,\dag}}
\newcommand{\affiliations}[1]{\affiliation{#1}}

\author{
    Yushan Liu\textsuperscript{\rm 1,2}\internshipwork,
    Tianxiong Lv\textsuperscript{\rm 2}\correspondingauthor\projectleader,
    Bohua Wang\textsuperscript{\rm 2},
    Hangqi Fan\textsuperscript{\rm 2},
    Chenxu Zhao\textsuperscript{\rm 2},
    He Zheng\textsuperscript{\rm 2},\\
    Xuchang Zhong\textsuperscript{\rm 2}\internshipwork,
    Yifan Xie\textsuperscript{\rm 1},
    Congyang Zhao\textsuperscript{\rm 2},
    Zhihao Liao\textsuperscript{\rm 2},
    Leigang Luo\textsuperscript{\rm 2},\\
    Yang Cai\textsuperscript{\rm 2},
    Xiao-Ping Zhang\textsuperscript{\rm 1},
    Wenbo Ding\textsuperscript{\rm 1}\correspondingauthor
}
\affiliations{
    \textsuperscript{\rm 1}Tsinghua University \quad
    \textsuperscript{\rm 2}AMap, Alibaba Group
}

\contribution[*]{Work done during internship at AMap, Alibaba Group}
\contributionbreak
\contribution[\dag]{Corresponding authors}
\contribution[\ddag]{Project leader}

\renewcommand{\authorfont}{\fontsize{11}{13}\selectfont}
\renewcommand{\affiliationfont}{\fontsize{10.5}{12.5}\selectfont}
\renewcommand{\contributionfont}{\fontsize{10}{12}\selectfont}

\abstract{

Frozen perception foundation models encode rich geometric, semantic, and dynamic knowledge. Yet narrow conditioning interfaces may attenuate task-relevant cues, while static fusion cannot adjust expert contributions to each scene. We cast this challenge as the \textit{prior-to-plan} transfer problem and introduce \textbf{PerceptDrive}, a perception prior world-action modeling framework with adaptive expert routing. PerceptDrive feeds teacher-distilled priors from a frozen, driving-adapted provider and dense observation latents from a frozen self-supervised video encoder into a trainable expert-routed world-action model. Expert-specific query branches process these signals, while a prior-retention objective anchors each branch to its prior. A router predicts soft gates from a shared scene representation and combines the expert conditions before trajectory generation.
During training, privileged rule-based sub-metric estimates for branch-specific trajectory drafts provide soft-gate distillation targets. The predicted action-free future latent conditions a flow-matching actor. At inference, privileged components are absent; with one front-facing camera, PerceptDrive generates one trajectory per planning step without test-time scoring, reranking, or search.
Experiments show that PerceptDrive achieves state-of-the-art performance with \textbf{90.4} PDMS on NAVSIM v1 and \textbf{90.2} EPDMS on NAVSIM v2, 
outperforming existing methods. 
Ablations confirm complementary gains from prior retention and scene-conditioned routing, alongside differential reliance on the three priors. These results demonstrate that preserving and adaptively routing perception priors improves direct planning without test-time candidate selection.

\bigskip
\textbf{Date:} July 20, 2026

\textbf{Correspondence:} \email{tianxiong.ltx@alibaba-inc.com}, \email{ding.wenbo@sz.tsinghua.edu.cn}

}

\begin{document}
\maketitle

\section{Introduction}

\begin{figure}[!tp]
  \centering
  \includegraphics[width=\textwidth]{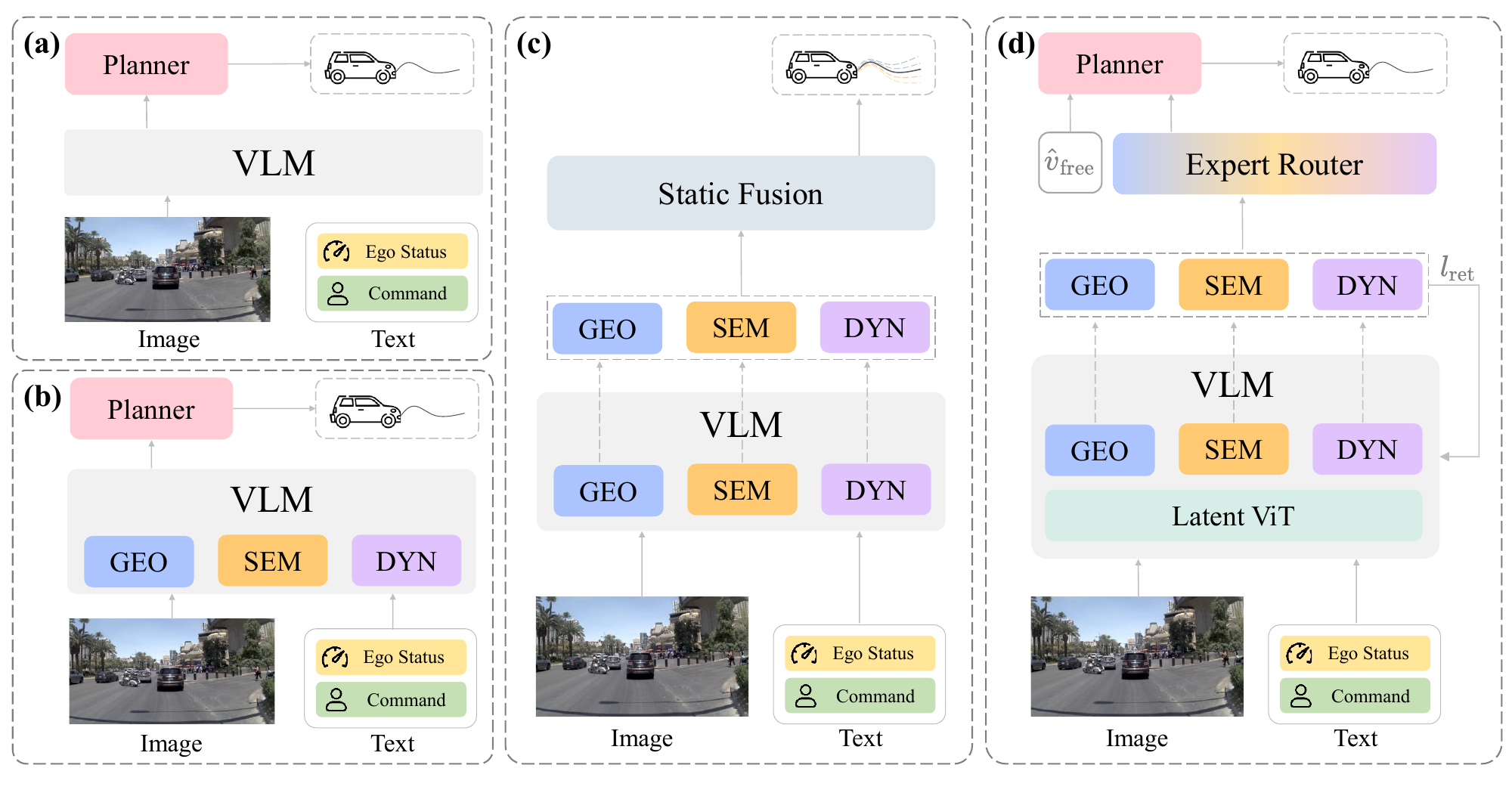}
  \caption{\textbf{Comparison of VLM-based planning paradigms.} (a) Direct VLM conditioning: knowledge stays implicit in a single narrow interface, where task-relevant cues are attenuated. (b) Expert priors inside the VLM: without explicit separation, no objective specifies which prior survives compression. (c) Separated expert representations under static fusion: scene-invariant weights cannot adapt each prior's contribution to the current scene. (d) \textbf{PerceptDrive (Ours)}: per-branch retention anchors each compressed readout to its prior, and a scene-conditioned router reweights the expert conditions per scene, yielding one trajectory under self-predicted future conditioning without test-time scoring or reranking.}
  \label{fig:compare}
\end{figure}

End-to-end autonomous driving maps onboard observations, ego states, and navigation commands directly to future trajectories. Early approaches convert sensor observations into planning-oriented representations~\citep{transfuser,uniad}; vision-language-action (VLA) models add language-conditioned semantics and reasoning~\citep{senna,autovla,futurevla}; and latent world models couple scene evolution with trajectory generation~\citep{law,drivinggpt,drivewam}. These developments have made increasingly rich geometric, semantic, and dynamic representations available to end-to-end planners.

However, two mismatches limit the transfer of this knowledge into continuous trajectory generation: First, query-based planners compress heterogeneous signals into a compact set of conditioning tokens. Standard imitation objectives supervise the final trajectory but do not explicitly specify which source-specific information should survive compression; scenario-dependent cues may therefore be attenuated. Second, static fusion uses scene-invariant weights and cannot adapt the relative contribution of each prior to the current scene~\citep{vlmad,driveworldvla,coworldvla}.

Existing methods address different aspects of the \emph{prior-to-plan transfer} problem by: injecting foundation knowledge into upstream representations through supervision, distillation, or expert tokens~\citep{vlmad,drivekd}; supplying predictive context via latent world modeling~\citep{latentwam,drivefuture}; or adapting computation and action selection to the scene with mixture-of-experts (MoE) or evaluator-based planners~\citep{drivemoe,gtrs}. These approaches, however, typically target different stages of the planning pipeline: scene-adaptive MoE routers generally route computation among parameter experts rather than modulate reliance on distinct frozen knowledge sources, and evaluator-based methods require candidate scoring or search at test time. What remains unresolved is a planner-side mechanism that both preserves source-specific information through query compression and adapts the resulting expert conditions to each scene before direct trajectory generation.

We introduce \textbf{PerceptDrive}, a perception prior world-action modeling framework with adaptive expert routing (Fig.~\ref{fig:compare}). A frozen, driving-adapted perception provider supplies teacher-distilled geometric, semantic, and dynamic priors, while a frozen self-supervised video encoder supplies dense observation latents. A trainable expert-routed world-action model integrates both streams, with its experts implemented as prior-specific conditioning branches rather than sparse parameter subnetworks. Expert-specific query branches compress these signals, and a per-branch prior-retention objective anchors each branch to its corresponding prior, preserving source-specific information through the query bottleneck. A router predicts soft gates from a shared scene representation and combines the expert conditions before trajectory generation, allowing their relative contributions to vary across scenes. During training, privileged rule-based sub-metric estimates for branch-specific trajectory drafts provide soft-gate distillation targets. The predicted action-free future latent then conditions a flow-matching actor. At inference, privileged components are absent, and PerceptDrive generates one trajectory per planning step without test-time scoring, reranking, or search.

Experiments show that PerceptDrive achieves state-of-the-art performance: PDMS of \textbf{90.4} on NAVSIM v1~\citep{navsim} and EPDMS of \textbf{90.2} on NAVSIM v2~\citep{navsimv2}, outperforming existing methods. Ablations support cumulative gains from prior retention and metric-distilled routing, favor scene-conditioned fusion of expert conditions over static weighting and trajectory averaging, and reveal differential reliance on the three priors. Our contributions are as follows:
\begin{itemize}
  \item We formulate \textit{prior-to-plan} transfer as a representational interface problem between a frozen, heterogeneous perception provider and a continuous world-action planner, decomposing it into compression- and scene-allocation-level mismatches; \textbf{PerceptDrive} is an expert-routed world-action framework targeting both.
  \item We introduce two coupled planner-side mechanisms: per-branch prior retention, which reconstructs stop-gradient expert representations from compressed readouts to anchor branches and counteract collapse; and metric-distilled expert routing, which scores training-only branch drafts with privileged sub-metrics and distills them into the gates, keeping inference single-trajectory.
  \item PerceptDrive sets a new state-of-the-art (\textbf{90.4} PDMS, \textbf{90.2} EPDMS); ablations support the cumulative component gains, the advantage of conditioning-level gating over static weighting and trajectory averaging, and the complementary roles of the three priors.
\end{itemize}

\section{Related Work}

\paragraph{Foundation priors for driving}
Foundation knowledge enters end-to-end driving mainly through two routes. One integrates language-aligned representations or their high-level decisions into the policy to support semantic reasoning, instruction following, and trajectory generation~\citep{recogdrive,autovla,senna}. The other transfers structured knowledge into driving representations via reasoning and action supervision, predictive video pretraining, or multi-teacher distillation~\citep{vlmad,drivejepa,drivekd}; recent methods further compress frozen visual features with learnable queries, organize geometric, semantic, and dynamic knowledge as structured perception or expert tokens~\citep{frostdrive,perceptwam,coworldvla}, or decouple semantic and spatial streams inside the policy~\citep{ssvla}. These paradigms enrich the representations available before action decoding, but they primarily target prior acquisition or upstream representation learning. PerceptDrive instead targets the compression interface through per-branch prior retention, which anchors each compressed branch readout to its corresponding prior.

\paragraph{Latent world-action models}
Latent world-action modeling links temporal prediction with continuous planning at several levels: predicting future latents or jointly modeling visual and action tokens as self-supervision for planning representations~\citep{law,drivinggpt,oawam}; coupling scene evolution with trajectory generation through future-image prediction, video-generation latents, action-conditioned imagination, alternating future--action prediction, or bidirectional visual--action cross-conditioning~\citep{drivevlaw0,drivelaw,driveworldvla,uniworldvla,forgedrive}; and conditioning the planner directly on predicted future latents with additional geometric or generative supervision~\citep{latentwam,drivedreamerpolicy,drivewam,drivefuture}; recent variants further couple ego--environment co-evolution with proactive planning~\citep{prodrive} or bridge latent style dynamics into downstream planning behavior~\citep{plans}. Future prediction supplies complementary temporal context for continuous planning, but does not by itself specify how heterogeneous frozen priors should be retained after compression or weighted across scenes. PerceptDrive combines conditioning on a predicted action-free future latent with per-branch prior retention and scene-conditioned routing.

\paragraph{Mixture-of-experts and routed planners}
MoE models allocate computation across experts through input-conditioned gates~\citep{moe,sparsemoe}. In driving, one line routes visual inputs or network modules by scene context or skill; another organizes semantic, geometric, and dynamic knowledge as expert conditions and fuses the resulting representations or trajectories~\citep{drivemoe,coworldvla}. Module routing, however, does not itself preserve information from a particular knowledge source, and knowledge-expert fusion may rely on scene-invariant weights or act only after complete trajectories are formed. A related family selects actions from fixed trajectory vocabularies, evaluates candidates with distilled sub-metric scorers or world-model rewards~\citep{hydramdp,vadv2,world4drive,gtrs,wote}, or refines a unified candidate set with scene-adaptive diffusion~\citep{unicand}; this aligns planning with quality evaluation but bounds coverage by the candidate set and requires test-time candidate evaluation. PerceptDrive instead couples branch-level retention with dense, conditioning-level gates distilled from privileged sub-metrics during training only, fusing expert conditions before generation and keeping inference a direct single trajectory.

\section{Method}
\label{sec:method}

\begin{figure*}[!tp]
  \centering
  \includegraphics[width=\textwidth]{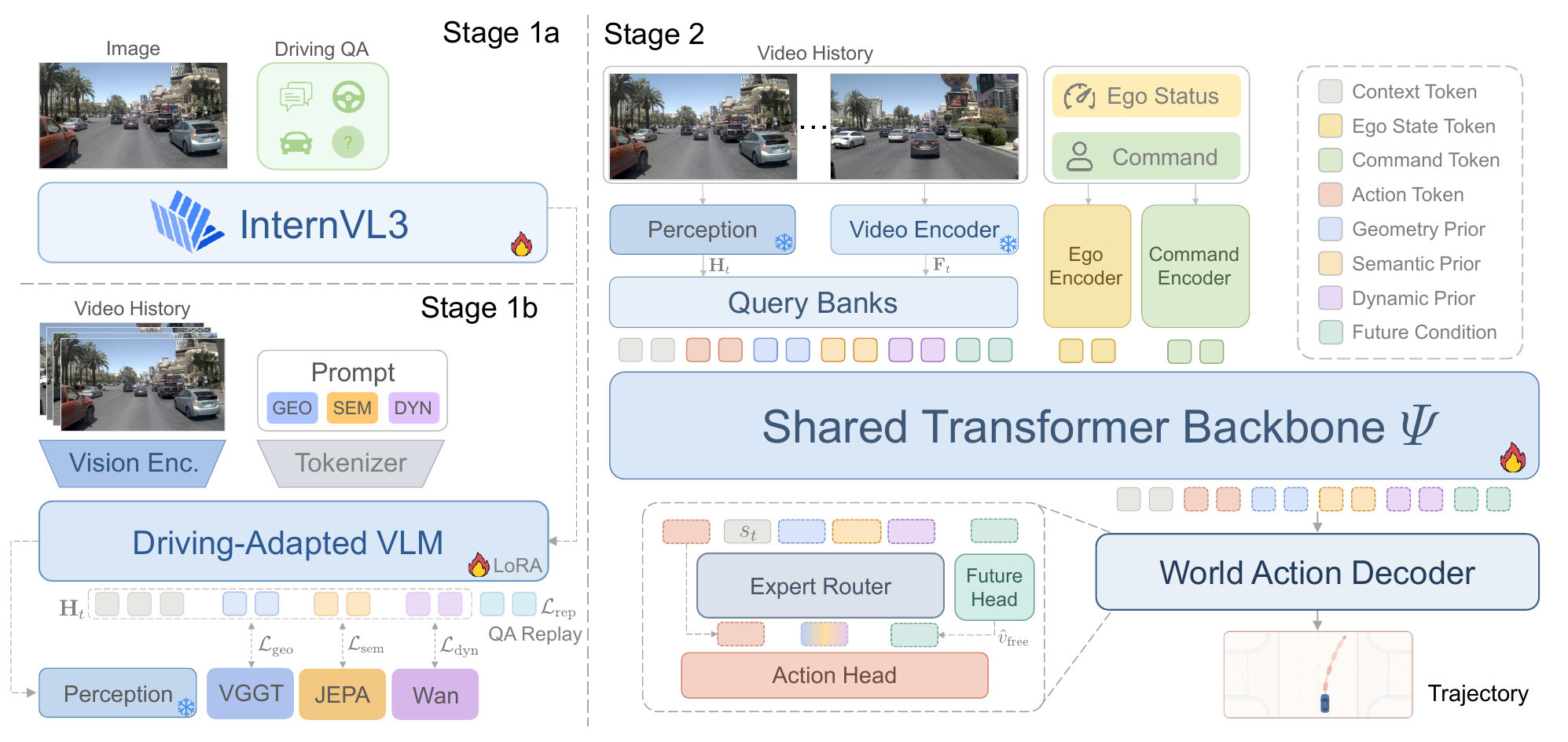}
  \caption{\textbf{Overview of PerceptDrive}. Stages~1a/1b build the frozen provider: driving-QA adaptation of InternVL3, then multi-teacher distillation of the \texttt{[GEO]}/\texttt{[SEM]}/\texttt{[DYN]} priors from frozen VGGT, V-JEPA~2, and Wan~2.1 teachers with QA replay. Stage~2 trains the world-action model: query banks compress $\mathbf H_t$ and $\mathbf F_t$, the backbone $\varPsi$ fuses them with ego state and command, and the router, future head, and action head generate one trajectory.}
  \label{fig:method}
\end{figure*}

\textbf{PerceptDrive} separates feature acquisition from planning so that frozen priors can be preserved and recruited per scene at the provider-to-planner interface. As shown in Fig.~\ref{fig:method}, a frozen \emph{perception provider}---a driving-adapted VLM---maps the image history to high-level representations $\mathbf{H}_t$ carrying geometric, semantic, and dynamic \emph{expert priors}, and an off-the-shelf frozen self-supervised video encoder supplies dense \emph{observation latents} $\mathbf{F}_t$. The two streams are complementary in granularity: the provider exposes compact, teacher-aligned expert slots, while the encoder preserves the dense spatio-temporal detail that compact slots necessarily discard. A trainable expert-routed world-action model (WAM) compresses both streams through query banks, anchors each expert branch to its designated prior with a retention objective, and predicts soft gates from a shared scene state to combine the expert conditions per scene. Conditioned on the gated combination and a self-predicted future latent, a flow-matching actor generates a single trajectory; the privileged supervision used during training is absent at inference.

\subsection{Problem Formulation}

We consider end-to-end planning from a single front-facing camera. At time $t$ the agent observes $o_t=(\mathcal{I}_t,\mathbf{e}_t,c_t)$: the front-view image history, per-frame ego states (position, heading, velocity, acceleration, yaw rate), and a navigation command $c_t\in\mathcal{C}$. From this observation, the agent predicts a future ego-centric BEV trajectory $\mathbf{a}=\{(x_i,y_i,\psi_i)\}_{i=1}^{H}\in\mathbb{R}^{H\times3}$, whose $H{=}8$ waypoints cover $4$\,s at $0.5$\,s intervals (inputs and normalization in Appendix~\ref{app:implementation}).

\subsection{Frozen Perception-Provider Construction}
\label{sec:provider}

The perception provider is a driving-adapted VLM whose registered expert slots carry geometric, semantic, and dynamic priors, exposed to the planner as $\mathbf{H}_t$. It is built in two steps before WAM optimization (Stages~1a/1b in Fig.~\ref{fig:method})---driving-QA adaptation, then multi-teacher expert distillation---which keeps the retention targets stationary; the low-level stream is not involved in this construction: the video encoder remains frozen and supplies $\mathbf{F}_t$ to the WAM directly.

The first step adapts InternVL3-2B~\citep{internvl3} to driving language and reasoning by full fine-tuning on 1,398,858 cleaned and deduplicated training samples from DriveLM, LingoQA, DriveQA, NuScenes-QA, and Reason2Drive~\citep{drivelm,lingoqa,driveqa,nuscenesqa,reason2drive}, covering scene understanding, road structure, traffic rules, navigation intent, and decision explanation; each sample is a task-prompted exchange with next-token cross-entropy, evaluated only on the answer segment: 
\begin{equation}
\mathcal{L}_{\mathrm{LM}}=-\sum\nolimits_{j\in\mathcal{Y}}\log p(y_j\mid y_{<j},x),
\end{equation}
where $x$ contains the prompt and visual context and $\mathcal{Y}$ indexes answer tokens.

Language supervision alone does not constrain the geometric structure, spatio-temporal semantics, and future dynamics that planning requires. The second step therefore appends three groups of registered expert tokens $\{\texttt{[GEO]},\texttt{[SEM]},\texttt{[DYN]}\}$ and distills complementary signals from frozen foundation models into their last-layer states---the \emph{expert slots}; for prior type $c$, a prior-specific head maps the slot state to the teacher space, $z^{(c)}=\phi_c\big(h^{(c)}\big)$. For geometry, a frozen VGGT teacher~\citep{vggt} supplies $F^{\mathrm{geo}}$, aligned in direction and magnitude:
\begin{equation}\label{eq:geo}
\mathcal{L}_{\mathrm{geo}}=\big(1-\cos(z^{\mathrm{geo}},F^{\mathrm{geo}})\big)+\tfrac{1}{2}\,\big\lVert z^{\mathrm{geo}}-F^{\mathrm{geo}}\big\rVert_2^2.
\end{equation}
For spatio-temporal semantics, a frozen V-JEPA~2 ViT-g teacher~\citep{vjepa2} provides context-predictive targets under multi-block masking at ratio $0.7$: a lightweight predictor reconstructs the teacher features at masked positions from the visible slots and mask pattern:
\begin{equation}\label{eq:sem}
\mathcal{L}_{\mathrm{sem}}=\big\lVert \mathrm{Pred}(z^{\mathrm{sem}}_{\mathrm{vis}},M)-F^{\mathrm{sem}}_{\mathrm{mask}}\big\rVert_2^2.
\end{equation}
For dynamics, a frozen Wan~2.1 video-diffusion teacher~\citep{wan} transfers its generative velocity field: its frozen VAE encodes the ground-truth future clip into $w$; for a sampled level $\sigma\in[0,1]$ and noise $\varepsilon$, $x_\sigma=(1-\sigma)w+\sigma\varepsilon$, and a trainable cross-attention bridge conditions the frozen denoiser $D$ on the dynamic slot:
\begin{equation}\label{eq:dyn}
\mathcal{L}_{\mathrm{dyn}}=\big\lVert D(x_\sigma,\sigma;\,z^{\mathrm{dyn}})-(\varepsilon-w)\big\rVert_2^2.
\end{equation}

All teachers remain frozen; gradients reach only the expert-token embeddings, the projection heads, the bridge, and the VLM LoRA adapters~\citep{lora}; a QA-replay mini-batch contributes $\mathcal{L}_{\mathrm{rep}}$ to preserve driving-language alignment, and the dynamic objective is introduced after geometric and context-predictive alignment. The combined provider loss is
\begin{equation}\label{eq:prior}
\mathcal{L}_{\mathrm{prior}}=\lambda_{\mathrm{geo}}\mathcal{L}_{\mathrm{geo}}+\lambda_{\mathrm{sem}}\mathcal{L}_{\mathrm{sem}}+\lambda_{\mathrm{dyn}}\mathcal{L}_{\mathrm{dyn}}+\lambda_{\mathrm{rep}}\mathcal{L}_{\mathrm{rep}},
\end{equation}
with the four $\lambda$ coefficients weighting the terms. After distillation the LoRA weights are merged and the VLM is frozen.

By contrast, a pretrained V-JEPA~2-L encoder supplies the low-level stream without driving-specific fine-tuning. It maps the same four-frame history to dense observation latents $\mathbf{F}_t$. The encoder also processes the corresponding future frames to produce the target $v_{\mathrm{gt}}$ used to supervise the future head (Sec.~\ref{sec:wam}). Because the current latents and future targets are produced by the same frozen encoder, the future head is trained within a common, stationary representation space. During WAM training and inference, both the adapted VLM and V-JEPA encoder remain frozen and provide the high- and low-level visual conditions, respectively. Implementation details for the encoder configuration and latent normalization are provided in Appendix~\ref{app:implementation}.

\subsection{Prior-Preserving World-Action Planning with Adaptive Expert Routing}
\label{sec:wam}

Two requirements shape the WAM: compressing the frozen input streams $\mathbf{H}_t$ and $\mathbf{F}_t$ without discarding planning-relevant information, and recruiting that information in a scene-dependent manner. Both are projected to a common width and concatenated into a perception pool $\mathcal{P}_t$, read through shared cross-attention by learnable query banks---global-context, expert-agnostic action, and temporal (video) banks, plus three \emph{expert-branch} banks $Q_c$, $c\in\{\mathrm{geo},\mathrm{sem},\mathrm{dyn}\}$, each dedicated to one prior and anchored to it by the retention objective in Eq.~\eqref{eq:ret}. A bidirectional Transformer backbone $\varPsi$ fuses the readouts with the ego-state and command embeddings, yielding per-expert conditions $\hat{c}_{c}$, the condition $\hat{c}_{\mathrm{vid}}$ for the future head, the condition $\hat{c}_{\mathrm{act}}$---joined with the gated expert mixture $\hat{c}_{\mathrm{exp}}$ (Eq.~\eqref{eq:route})---for the action head, and an attention-pooled scene vector $s_t$ for metric supervision and routing (sequence layout in Appendix~\ref{app:implementation}).

The query interface constitutes an information bottleneck: The query banks compress $\mathcal{P}_t$ into branch readouts $C_c$, after which $\varPsi$ operates only on these representations and no longer accesses the original perception pool. The base objectives do not specify which expert-specific information each $C_c$ should preserve. Because all branches are optimized under the same imitation objective, they may learn redundant representations. We therefore attach a lightweight retention probe $\rho_c$ to each branch to reconstruct the detached expert-slot target $\bar{h}^{c}$ from the mean-pooled readout $\bar{C}_c$ (targets in Appendix~\ref{app:implementation}), reusing the alignment of Eq.~\eqref{eq:geo}:
\begin{equation}\label{eq:ret}
\mathcal{L}_{\mathrm{ret}}=\sum_{c}\big[1-\cos(\rho_c(\bar{C}_c),\bar{h}^{c})+\tfrac{1}{2}\lVert\rho_c(\bar{C}_c)-\bar{h}^{c}\rVert_2^2\big].
\end{equation}
The probes are train-only; because each branch reconstructs a distinct target, their gradients pull each readout toward its designated prior, counteracting collapse (representation-level verification in Table~\ref{tab:retention_diag}). Anchoring compressed readouts, rather than passing the slot states forward directly, allows each branch to integrate its prior with the full perception pool---including the dense observation latents---while remaining specialized; restricting each branch's access to its designated slots underperforms this design (Table~\ref{tab:abl_design}).

The relative importance of the expert conditions varies across scenes. We therefore mix the experts softly at the \emph{conditioning} level: a two-layer MLP router $g_r$ predicts gates from $s_t$ and combines the expert conditions,
\begin{equation}\label{eq:route}
\boldsymbol\alpha=\operatorname{softmax}\big(g_r(s_t)\big)\in\Delta^{2},\qquad
\hat{c}_{\mathrm{exp}}=\sum_{c}\alpha_c\,\hat{c}_{c}.
\end{equation}
The gates lie on $\Delta^{2}$, the probability simplex over the three experts: all three experts remain active, and $\hat{c}_{\mathrm{exp}}$ is a dense, scene-weighted combination of their retention-anchored conditions. Since $\boldsymbol\alpha$ depends only on $s_t$, gating is feed-forward and precedes generation.

Generation couples a future head with a flow-matching actor. The future head $\mathrm{FP}$ predicts the next $2$\,s of frozen-encoder latents in two modes: an action-free latent $\hat{v}_{\mathrm{free}}$ that conditions the actor, and an action-conditioned latent $\hat{v}(\mathbf{a})$ for the auxiliary branch, both $\ell_1$-regressed against the frozen-encoder target $v_{\mathrm{gt}}$. The actor's velocity network $v_\theta$~\citep{flowmatching,rectifiedflow}, conditioned on $[\hat{c}_{\mathrm{act}};\hat{c}_{\mathrm{exp}}]$ and, through a zero-initialized cross-attention residual, on the detached $\hat{v}_{\mathrm{free}}$, is trained with the standard flow-matching objective $\mathcal{L}_{\mathrm{act}}$ and integrates from base noise to one trajectory at inference (formulations in Appendix~\ref{app:training_details}; wiring ablated in Table~\ref{tab:abl_design}).

Imitation and future reconstruction provide no assessment of an action's safety, compliance, comfort, or progress; privileged rule-based sub-metrics therefore supervise the WAM at two points during training. First, an auxiliary regressor $g_\omega$ predicts a $K$-dimensional quality vector from $s_t$, an action, and its imagined outcome, supervised on an offline pool of privileged-scored trajectories with $k$-nearest-neighbor targets for in-batch predictions; its action and future inputs are detached, so $\mathcal{L}_{\mathrm{aux}}$ reaches the backbone only through $s_t$. Second, because an end-to-end gate need not align with driving quality, the same scores are distilled into the router: three one-step \emph{branch drafts}, formed under one-hot gating by reusing the flow interpolation of the action loss, are scored by the same pool interpolation and converted at temperature $T_r$ into a target gate $\alpha^{\ast}$ supervising $g_r$ via the cross-entropy $\mathcal{L}_{\mathrm{route}}$. Both supervision pathways are restricted to training and are omitted at inference. Appendix~\ref{app:training_details} defines their objectives and details the construction of the offline trajectory pool and the one-step branch-conditioned trajectory estimates; these surrogates only construct routing supervision on navtrain, and all reported PDMS and EPDMS values are computed directly by the corresponding official NAVSIM evaluator on navtest or navhard. This design amortizes test-time trajectory scoring into training-time gate supervision: the evaluator's knowledge is distilled into the gates, while inference remains a single feed-forward plan.

\subsection{Optimization and Single-Trajectory Inference}

Training proceeds in two stages: the provider is constructed first (driving-QA adaptation with $\mathcal L_{\mathrm{LM}}$, multi-teacher distillation with $\mathcal L_{\mathrm{prior}}$, then frozen), after which all WAM modules are optimized in two phases---a \emph{warmup} with $\mathcal L_{\mathrm{act}}$ and $\mathcal L_{\mathrm{route}}$ disabled and gating held uniform, then a \emph{joint} phase minimizing
\begin{equation}\label{eq:total}
\mathcal{L}=\mathcal{L}_{\mathrm{act}}+\lambda_f\,\mathcal{L}_{\mathrm{fut}}+\mathcal{L}_{\mathrm{aux}}+\lambda_{\mathrm{ret}}\,\mathcal{L}_{\mathrm{ret}}+\lambda_r\,\mathcal{L}_{\mathrm{route}},
\end{equation}
where the action and auxiliary terms have unit weight and $\lambda_f$, $\lambda_{\mathrm{ret}}$, $\lambda_r$ weight the remaining terms; phase lengths, weights, temperature, and the trajectory pool are given in Appendix~\ref{app:training_details}. Gradient routing is deliberate: $\mathcal{L}_{\mathrm{act}}$ does not update the future head ($\hat v_{\mathrm{free}}$ is detached), $\mathcal{L}_{\mathrm{aux}}$ and $\mathcal{L}_{\mathrm{route}}$ reach the shared backbone only through $s_t$, retention updates only the probes and branch readouts, and branch drafts are stop-gradient (full routes in Appendix~\ref{app:training_details}).

At inference, the provider encodes the observation, the router produces $\boldsymbol\alpha$ in one feed-forward pass, the action-free path predicts $\hat v_{\mathrm{free}}$, and the flow actor yields one trajectory by multi-step Euler integration. All privileged components are absent; the expert branches condition a single model rather than form a candidate set, so no candidate scoring, reranking, or test-time search is required.

\section{Experiments}

\begin{figure*}[!tp]
  \centering
  \includegraphics[width=\textwidth]{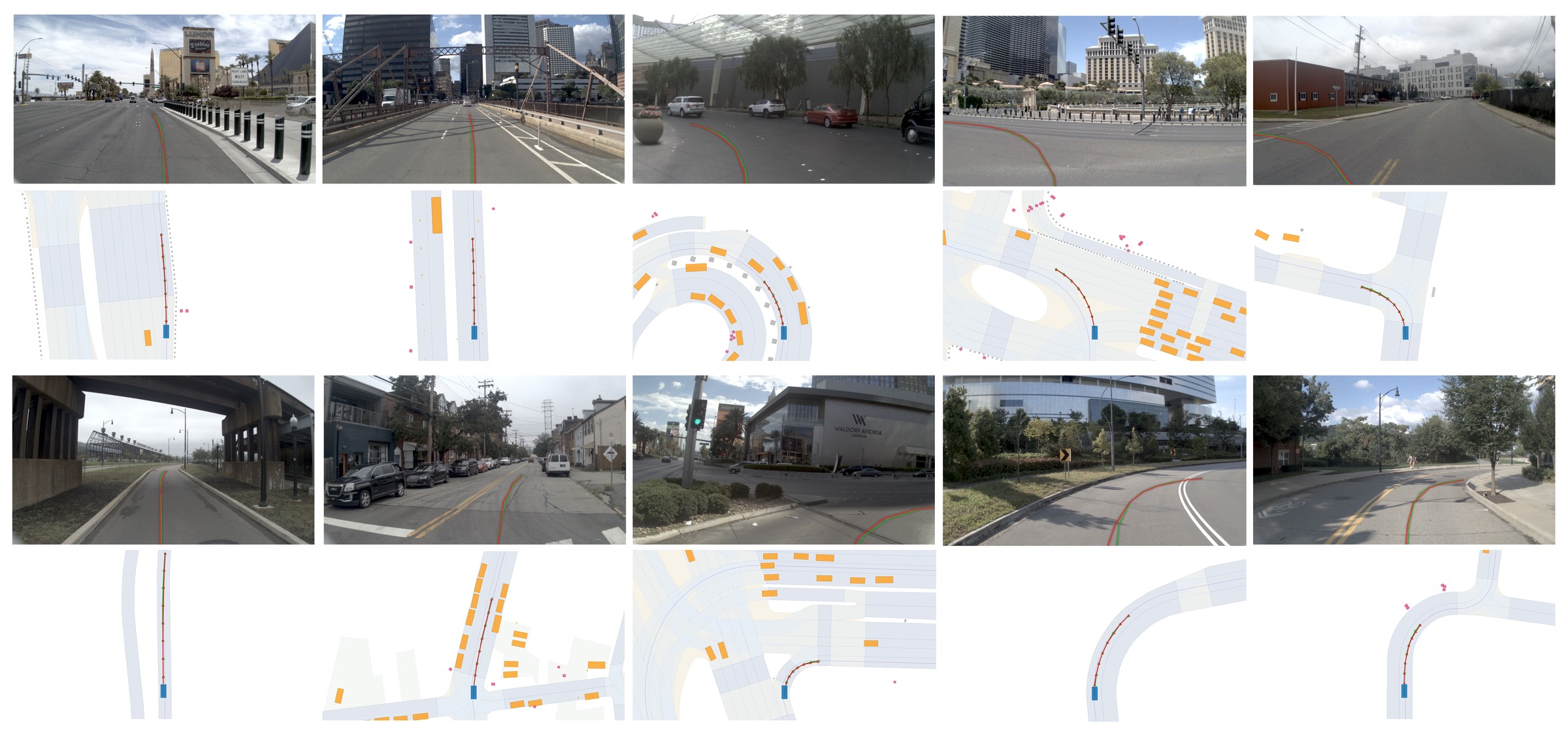}
  \caption{\textbf{Qualitative results on NAVSIM navtest.} Each example pairs the front-view observation (top) with its BEV context and the generated trajectory (bottom). \trajectorylegend}
  \label{fig:qualitative_navsim}
\end{figure*}

\subsection{Experimental Setup}
We train the WAM on NAVSIM navtrain and evaluate PerceptDrive under three protocols~\citep{navsim,navsimv2}: NAVSIM v1 PDM Score (PDMS) on navtest, NAVSIM v2 Extended PDM Score (EPDMS) on navtest, and NAVSIM v2 two-stage simulation on navhard. All offline supervision pools are constructed exclusively from navtrain. Appendix~\ref{app:evaluation_protocols} details the metrics and protocols; Appendix~\ref{app:implementation} gives implementation and inference settings. Appendix~\ref{app:cost} reports the training cost and parameter profile.

\subsection{Results}
Tables~\ref{tab:pdms_navtest} and~\ref{tab:epdms_navtest} compare direct-planning results on navtest. With one front-facing camera and no candidate scoring or test-time optimization, PerceptDrive achieves state-of-the-art results: \textbf{90.4} PDMS under NAVSIM v1, and \textbf{90.2} EPDMS under NAVSIM v2 evaluation, both above the state-of-the-art baseline. Across three training seeds the full model is stable at $90.4{\pm}0.05$ PDMS and $90.2{\pm}0.11$ EPDMS, with evaluation-seed standard deviation at most $0.05$ for all variants, and per-scene paired bootstrap intervals separate it from every internal ablation (Table~\ref{tab:multiseed}). PerceptDrive also achieves 34.5 EPDMS on NAVSIM v2 navhard (Table~\ref{tab:navhard}). Figure~\ref{fig:qualitative_navsim} shows qualitative visualization in which the generated trajectory closely follows the logged path across intersections, turns, and dense traffic; Appendix~\ref{app:visualizations} provides more examples.

\begin{table*}[!t]
\centering
\small
\setlength{\tabcolsep}{8pt}
\caption{\textbf{Comparison with state-of-the-art methods on NAVSIM v1 navtest.} $N{\times}$C denotes $N$ camera views and $+$L denotes LiDAR; ``--'' indicates unavailable or non-comparable sub-scores. \textbf{Bold}/\underline{underline} mark the best/second values among the listed non-human methods. The Comf.\ column is saturated (99.3--100.0) and carries little discriminative signal.}
\label{tab:pdms_navtest}
\begin{tabular}{lcrrrrr>{\columncolor{pdmsgray}}r}
\toprule
\textbf{Method} & \textbf{Sensors} & \textbf{NC$\uparrow$} & \textbf{DAC$\uparrow$} & \textbf{EP$\uparrow$} & \textbf{TTC$\uparrow$} & \textbf{Comf.$\uparrow$} & \textbf{PDMS$\uparrow$} \\
\midrule
Human            & --              & 100.0 & 100.0 & 87.5 & 100.0 & 99.9 & 94.8 \\
\midrule
\rowcolor{groupgray}\multicolumn{8}{l}{\textit{VLA-based Methods}} \\
ReCogDrive~\citep[arXiv v1]{recogdrive} & $1{\times}$C & 98.2 & 97.8 & \underline{83.5} & 95.2 & 99.8 & 89.6 \\
AutoVLA~\citep{autovla}             & $3{\times}$C    & 98.4 & 95.6 & 81.9 & \textbf{98.0} & \underline{99.9} & 89.1 \\
DriveVLA-W0~\citep{drivevlaw0}         & $1{\times}$C    & 98.7 & \textbf{99.1} & 83.3 & 95.3 & 99.3 & \underline{90.2} \\
Uni-World-VLA~\citep{uniworldvla}       & $1{\times}$C    & 98.7 & 96.7 & 83.2 & 96.1 & \textbf{100.0} & 89.4 \\
\midrule
\rowcolor{groupgray}\multicolumn{8}{l}{\textit{World Model-based Methods}} \\
LAW~\citep{law}                  & $1{\times}$C    & 96.4 & 95.4 & 81.7 & 88.7 & \underline{99.9} & 84.6 \\
WoTE~\citep{wote}                & $3{\times}$C$+$L & 98.5 & 96.8 & 81.9 & 94.9 & \underline{99.9} & 88.3 \\
Epona~\citep{epona}                 & $1{\times}$C & 97.9 & 95.1 & 80.4 & 93.8 & \underline{99.9} & 86.2 \\
World4Drive~\citep{world4drive}         & $3{\times}$C    & 97.4 & 94.3 & 79.9 & 92.8 & \textbf{100.0} & 85.1 \\
SeerDrive~\citep{seerdrive}           & $3{\times}$C$+$L & 98.4 & 97.0 & 83.2 & 94.9 & \underline{99.9} & 88.9 \\
PWM~\citep{pwm}                 & $1{\times}$C    & 98.6 & 95.9 & 81.8 & 95.4 & \textbf{100.0} & 88.1 \\
DriveLaW~\citep{drivelaw}            & $1{\times}$C    & \underline{99.0} & 97.1 & 81.3 & \underline{96.7} & \textbf{100.0} & 89.1 \\
WorldDrive~\citep{worlddrive}          & $1{\times}$C    & 98.4 & 96.2 & 81.9 & 95.1 & \textbf{100.0} & 88.1 \\
DriveDreamer-Policy~\citep{drivedreamerpolicy} & $3{\times}$C    & 98.4 & 97.1 & \underline{83.5} & 95.1 & \textbf{100.0} & 89.2 \\
DriveWAM~\citep{drivewam}            & $1{\times}$C    & 98.3 & 98.1 & \textbf{84.3} & 95.2 & \textbf{100.0} & 90.1 \\
\midrule
\rowcolor{oursblue}\textbf{PerceptDrive (Ours)}  & $1{\times}$C & \textbf{99.2} & \underline{99.0} & 82.9 & \textbf{98.0} & \textbf{100.0} & \textbf{90.4} \\
\bottomrule
\end{tabular}
\end{table*}

\begin{table*}[!t]
\centering
\small
\setlength{\tabcolsep}{4pt}
\caption{\textbf{Comparison with state-of-the-art methods under NAVSIM v2 EPDMS navtest.} \textbf{Bold}/\underline{underline} mark the best/second values among the listed methods.}
\label{tab:epdms_navtest}
\begin{tabular}{lrrrrrrrrr>{\columncolor{pdmsgray}}r}
\toprule
\textbf{Method} & \textbf{NC$\uparrow$} & \textbf{DAC$\uparrow$} & \textbf{DDC$\uparrow$} & \textbf{TLC$\uparrow$} & \textbf{EP$\uparrow$} & \textbf{TTC$\uparrow$} & \textbf{LK$\uparrow$} & \textbf{HC$\uparrow$} & \textbf{EC$\uparrow$} & \textbf{EPDMS$\uparrow$} \\
\midrule
\rowcolor{groupgray}\multicolumn{11}{l}{\textit{VLA-based Methods}} \\
DriveVLA-W0~\citep{drivevlaw0}    & 98.5 & \textbf{99.1} & 98.0 & 99.7 & 86.4 & 98.1 & 93.2 & 97.9 & 58.9 & 86.1 \\
DriveWorld-VLA~\citep{driveworldvla} & 98.6 & \textbf{99.1} & \underline{99.6} & \underline{99.8} & 87.4 & 97.9 & 97.0 & 97.8 & 78.6 & 86.8 \\
HiST-VLA~\citep{histvla}            & \textbf{99.6} & \textbf{99.1} & \textbf{99.7} & \textbf{99.9} & \textbf{89.2} & \textbf{99.4} & \textbf{98.9} & \underline{98.4} & 66.0 & 88.6 \\
\midrule
\rowcolor{groupgray}\multicolumn{11}{l}{\textit{World Model-based Methods}} \\
Drive-JEPA~\citep{drivejepa}          & 98.4 & 98.6 & 99.1 & \underline{99.8} & \underline{88.4} & 97.8 & \underline{97.6} & 97.9 & 84.8 & 87.8 \\
Latent-WAM~\citep{latentwam}          & 98.1 & 97.3 & \underline{99.6} & \underline{99.8} & 87.7 & 97.3 & \underline{97.6} & 98.1 & \textbf{87.3} & 89.3 \\
DriveDreamer-Policy~\citep{drivedreamerpolicy} & 98.4 & 97.1 & 99.5 & \textbf{99.9} & 87.9 & 97.7 & \underline{97.6} & 98.3 & 79.4 & 88.7 \\
DriveFuture~\citep{drivefuture}         & \underline{98.8} & \textbf{99.1} & \underline{99.6} & \textbf{99.9} & 86.6 & \underline{98.4} & 96.4 & 98.3 & 74.8 & \underline{89.9} \\
IDOL~\citep{idol}                & \underline{98.8} & 97.6 & 99.5 & \underline{99.8} & 87.1 & 98.3 & 96.3 & 98.3 & 85.5 & 89.6 \\
GraphWorld~\citep{graphworld}          & 98.4 & \underline{98.8} & 99.1 & 99.1 & 85.9 & 97.9 & 96.0 & 97.8 & 74.6 & 89.5 \\
\midrule
\rowcolor{oursblue}\textbf{PerceptDrive (Ours)} & 98.6 & \underline{98.8} & 99.5 & \textbf{99.9} & 87.5 & 97.9 & 96.9 & \textbf{98.5} & \underline{86.3} & \textbf{90.2} \\
\bottomrule
\end{tabular}
\end{table*}

\FloatBarrier
\subsection{Ablation Studies}

\begin{figure*}[!tp]
  \includegraphics[width=\textwidth]{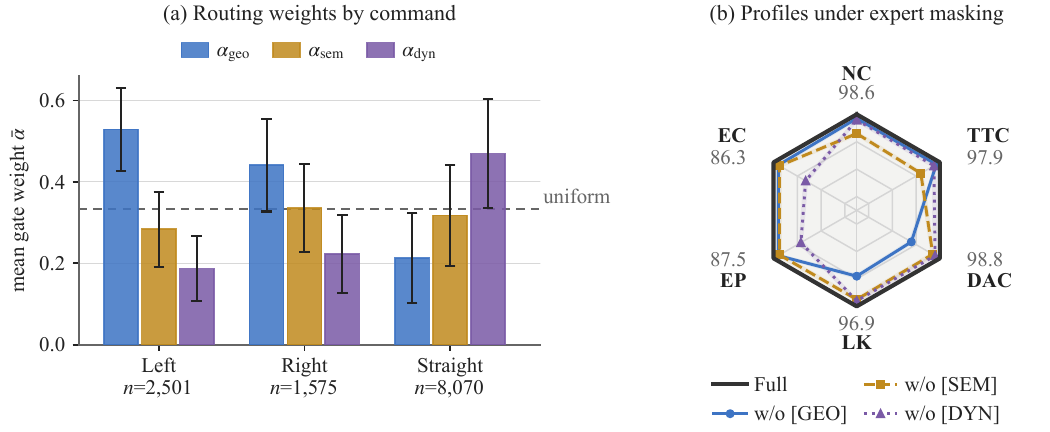}
  \caption{\textbf{Ablation Studies Visualization.} (a) Mean routing weights $\bar{\boldsymbol\alpha}$ by navigation command on \texttt{navtest} (whiskers: within-command std; $n$ below each command): gates shift toward the geometric expert under turns (Left/Right) and the dynamic expert under straight cruising, retaining within-command dispersion (mean std 0.11). (b) Sub-metric profiles under inference-time expert masking; each axis spans a uniform 7-point range below Full (rings: $-2/-4/-6$; shading: 95\% bootstrap bands). Masking degrades mainly the expert's own sector: \texttt{[SEM]} safety (NC, TTC), \texttt{[GEO]} road (DAC, LK), \texttt{[DYN]} progress/comfort (EP, EC).}
  \label{fig:ablation_viz}
\end{figure*}

\paragraph{Perception-prior acquisition}
Table~\ref{tab:abl_prior} shows that both the high-level VLM stream and multi-teacher prior distillation contribute to the complete provider. The inference-time token-removal diagnostic follows a consistent pattern: removing \texttt{[GEO]} most affects road geometry, \texttt{[SEM]} safety, and \texttt{[DYN]} progress and extended comfort. Figure~\ref{fig:ablation_viz}(b) visualizes this pattern as sub-metric profiles below the Full score, each expert's mask degrading mainly its own sector. Beyond inference-time masking, retraining the provider without the dynamic teacher (and re-optimizing the WAM) lowers EPDMS to 88.6 with the deficit again concentrated on EP and EC (Table~\ref{tab:abl_prior}); its persistence under retraining supports a causal contribution of the dynamic prior rather than an input-distribution artifact. Appendix~\ref{app:ablation_protocol} discusses the interpretive scope of the two variant families.

\paragraph{World-action model}
Table~\ref{tab:abl_wam} evaluates nested variants in the construction order of the planning pipeline: performance improves monotonically as future conditioning, metric-aware supervision, per-branch prior retention, and metric-distilled routing are introduced, and the complete configuration performs best under both evaluation protocols; variants without MR keep the three expert branches under uniform gating (nested design discussed in Appendix~\ref{app:ablation_protocol}). The ordering is preserved by all three training seeds (evaluation-seed std $\leq0.05$; Table~\ref{tab:multiseed}). The EC column localizes the extended-comfort gain---the most volatile sub-metric across published systems: most of it arises with future conditioning ($78.9\!\to\!84.0$), the remainder with the dynamic-prior pathway and routing. Table~\ref{tab:providers_mech} further disentangles the mechanisms from provider strength: prior retention and metric-distilled routing improve a QA-only provider by $+1.8$ EPDMS and the distilled provider by $+3.1$.

\begin{table}[!t]
\centering
\small
\setlength{\tabcolsep}{1.5pt}
\caption{\textbf{Ablation on the world-action model.} FC: future conditioning; MS: metric-aware supervision; PR: per-branch prior retention; MR: metric-distilled routing (without MR, gating is uniform). PDMS: NAVSIM v1 navtest; EPDMS: NAVSIM v2 navtest; subscripts: std over three training seeds (Table~\ref{tab:multiseed}); EC traces the extended-comfort gain. All variants share the training protocol.}
\label{tab:abl_wam}
\begin{tabular}{lccccrr>{\columncolor{pdmsgray}}r}
\toprule
\textbf{Variant} & \textbf{FC} & \textbf{MS} & \textbf{PR} & \textbf{MR} & \textbf{PDMS$\uparrow$} & \textbf{EC$\uparrow$} & \textbf{EPDMS$\uparrow$} \\
\midrule
Imitation only &  &  &  &  & 86.9$_{\pm.2}$ & 78.9 & 84.6$_{\pm.3}$ \\
+ Future conditioning & \checkmark &  &  &  & 87.5$_{\pm.2}$ & 84.0 & 85.4$_{\pm.1}$ \\
+ Metric supervision & \checkmark & \checkmark &  &  & 88.6$_{\pm.1}$ & 84.6 & 87.1$_{\pm.2}$ \\
+ Prior retention & \checkmark & \checkmark & \checkmark &  & 89.8$_{\pm.2}$ & 85.2 & 88.7$_{\pm.2}$ \\
\rowcolor{oursblue}PerceptDrive (full) & \checkmark & \checkmark & \checkmark & \checkmark & \textbf{90.4}$_{\pm.1}$ & \textbf{86.3} & \textbf{90.2}$_{\pm.1}$ \\
\bottomrule
\end{tabular}
\end{table}

\paragraph{Design choices}
Table~\ref{tab:abl_design} verifies the optimization and routing choices: standard initialization of the future-conditioning projection, single-phase training, and blocking the metric branch's gradient through $s_t$ all degrade both metrics, consistent with the intended gradient routes (Appendix~\ref{app:training_details}). The routing degradations are layered: trajectory-level averaging degrades performance the most (88.5 PDMS), supporting fusion in the conditioning space; a static learnable weighting matches uniform gating (89.7 vs.\ 89.8 PDMS), so the gain is not from learnable weights per se; and an end-to-end gate without distillation remains near-uniform (90.0 PDMS), leaving a clear gap to the full model---the benefit of scene-conditioned gating is realized largely through privileged metric distillation. A command-only gate reaches 89.0 EPDMS, 1.2 points below the full router, whereas removing the command from the router input reduces EPDMS by only 0.3; the routing benefit therefore derives mainly from scene features. In the retention block, shuffled branch--target pairing performs worse than removing the retention loss entirely, a shared concatenated target recovers only part of the gain, token-level targets match the pooled form, and restricting each bank's access to its designated slots slightly underperforms the anchored default---per-branch target assignment, not a generic reconstruction load, drives the gain (representation-level evidence in Table~\ref{tab:retention_diag}).

\paragraph{Routing behavior}
Figure~\ref{fig:ablation_viz}(a) reports the mean gate weights by navigation command: the geometric expert receives the largest weight under turns (Left/Right), while the dynamic expert dominates under straight cruising. The gates remain scene-dependent within commands---mean within-command std 0.11, a command-only predictor explaining only 38\% of the gate variance (Appendix~\ref{app:diagnostics}). Together with the command-only control in Table~\ref{tab:abl_design}, this pattern indicates scene-conditioned rather than merely command-conditioned gating (Eq.~\eqref{eq:route}), consistent with the masking sensitivities in Table~\ref{tab:abl_prior}.


\paragraph{Discussion}
PerceptDrive distills privileged rule-based sub-metrics into the representation and the gates, so part of the gain reflects evaluator alignment. We view this alignment as the intended mechanism, which amortizes a trusted evaluator into feed-forward planning and follows the lineage of scorer-distillation planners~\citep{hydramdp,gtrs}. Three observations indicate that it complements rather than substitutes for perceptual transfer. First, richer priors amplify the same mechanisms (+3.1 vs. +1.8 EPDMS; Table~\ref{tab:providers_mech}). Second, removing each prior degrades mainly its corresponding metric sector, and the dynamic-prior effect persists after retraining (Figure~\ref{fig:ablation_viz}(b); Table~\ref{tab:abl_prior}). Third, gate variance is dominated by scene features rather than command labels (Appendix~\ref{app:diagnostics}). The alignment target, however, is NAVSIM's specific scorer; transfer to differently weighted objectives is untested, and quantifying this sensitivity (e.g., scorer-subset distillation) is future work. All evaluations follow NAVSIM's non-reactive protocols; closed-loop evaluation is left open. The future head remains deterministic, and its action-conditioned predictions are extrapolations beyond demonstrated actions. Token masking probes reliance rather than isolated causal effects, with retraining verification limited to the dynamic prior.

\section{Conclusion}
We introduced \textbf{PerceptDrive}, a prior-to-plan interface between frozen foundation representations and continuous trajectory generation: per-branch retention anchors the expert readouts to geometric, semantic, and dynamic priors; a scene-conditioned router combines the expert conditions before generation; a predicted latent future conditions the flow actor; and privileged sub-metrics shape the representation and gates only during training, leaving single-camera, single-trajectory inference free of candidate scoring and reranking.

PerceptDrive attains state-of-the-art performance: \textbf{90.4} NAVSIM v1 PDMS and \textbf{90.2} NAVSIM v2 EPDMS on navtest. The ablations support the cumulative component gains, the complementary roles of the three priors, and the advantage of scene-conditioned conditioning-level gating over static weighting and trajectory averaging, largely realized through privileged metric distillation. Future work will focus on interactive closed-loop evaluation, multimodal futures, and a broader family of retained priors.

\clearpage

\bibliographystyle{plainnat}
\bibliography{ref}

\clearpage
\beginappendix

\setcounter{table}{0}\setcounter{figure}{0}\setcounter{equation}{0}
\renewcommand{\thetable}{A\arabic{table}}
\renewcommand{\thefigure}{A\arabic{figure}}
\renewcommand{\theequation}{A\arabic{equation}}

\section{Evaluation Protocols}
\label{app:evaluation_protocols}

\paragraph{Data splits}
The trainable WAM is optimized on NAVSIM navtrain and evaluated on navtest or navhard according to the protocol below. The perception provider is adapted separately using the driving-QA and teacher sources described in the main text, then frozen before WAM training. The offline trajectory-and-score pools used by the metric-aware branch contain navtrain samples only; neither navtest nor navhard contributes privileged supervision.

\paragraph{NAVSIM v1 PDMS on navtest}
PDMS combines no collision (NC), drivable-area compliance (DAC), ego progress (EP), time-to-collision (TTC), and comfort. Table~\ref{tab:pdms_navtest} compares published direct-planning methods that report navtest PDMS under this protocol, grouped into VLA-based and world-model-based families, together with the human-agent reference; for each method we quote the strongest configuration reported in the cited paper. Sensor coverage, external pretraining, proposal count, and inference budget differ across papers.

\paragraph{NAVSIM v2 EPDMS on navtest}
NAVSIM v2 introduces the Extended PDM Score (EPDMS), which extends PDMS with two new weighted metrics---lane keeping (LK) and extended comfort (EC)---two new multiplier metrics---driving-direction compliance (DDC) and traffic-light compliance (TLC)---and false-positive penalty filtering, under which a penalty is disabled when the human agent is also responsible for the violation. EPDMS aggregates the multiplier terms (NC, DAC, DDC, TLC) as a product and the weighted terms (EP, TTC, LK, HC, EC) as a weighted average, with history comfort (HC) in place of the v1 comfort term. Table~\ref{tab:epdms_navtest} compares reported results under this evaluation on navtest.

\paragraph{NAVSIM v2 two-stage pseudo-simulation on navhard}
The official public navhard evaluation uses two stages. Pre-generated follow-up observations provide the second-stage input rather than observations returned by a conventional interactive closed loop. Table~\ref{tab:navhard} reports sub-scores for both stages, and EPDMS is the final combined score.

\section{Implementation Details}
\label{app:implementation}

\paragraph{Inputs and trajectory target}
PerceptDrive uses one front-facing camera. Each observation contains four frames sampled at $2$\,Hz, an aligned ego-state history, and a high-level navigation command. The ego state records position, heading, velocity, acceleration, and yaw rate. The target contains eight ego-centric BEV waypoints at $0.5$\,s intervals, covering a $4$\,s horizon; each waypoint contains $(x,y,\psi)$. During WAM optimization, the waypoint channels are normalized by $(50,50,\pi)$ and converted back to physical coordinates for evaluation.

\paragraph{Frozen perception provider}

The high-level stream uses InternVL3-2B, first adapted on 1,398,858 cleaned and deduplicated driving-QA training samples (5,464 optimizer steps at effective batch size 256) and then aligned with frozen VGGT, V-JEPA~2 ViT-g, and Wan 2.1 teachers through two-layer prior-specific projection heads and a dynamic cross-attention bridge. A QA replay mini-batch is retained during multi-teacher distillation, and the dynamic teacher is introduced after geometric and context-predictive alignment. During distillation, gradients are restricted to the registered expert-token embeddings, these lightweight modules, and the VLM LoRA adapters; all foundation teachers remain frozen. After the LoRA weights are merged, the adapted VLM supplies $\mathbf H_t\in\mathbb R^{N_h\times1536}$ with

$N_h{=}1{,}048$ tokens (visual tokens plus the registered expert slots). A frozen V-JEPA~2-L encoder supplies $\mathbf F_t\in\mathbb R^{256\times2\times1024}$ for four frames with tubelet size 2; per-channel latent normalization is applied before the WAM. The adapted VLM constitutes the frozen perception provider; the off-the-shelf V-JEPA~2-L encoder separately supplies the low-level stream.

\paragraph{WAM capacity and optimization}

The WAM contains 421,481,483 trainable parameters (421M) and is optimized for 30k steps. Both frozen input streams are projected to width $d=1024$. The context, action, and video query banks contain 64, 8, and 32 tokens, respectively. Their readouts, together with the ego-state and command embeddings, enter the shared bidirectional Transformer as the sequence $[\,C_{\mathrm{ctx}},E_{\mathrm{ego}},E_{\mathrm{cmd}},C_{\mathrm{act}},C_{\mathrm{geo}},C_{\mathrm{sem}},C_{\mathrm{dyn}},C_{\mathrm{vid}}\,]$. The retention probes are two-layer MLPs; the expert-branch readouts are mean-pooled for retention against the slot targets $\bar h^{c}$---the arithmetic mean of slot-group tokens after the final VLM LayerNorm, detached from the provider---while the imagined future is attention-pooled for metric regression. The trainable WAM modules comprise the query banks, shared Transformer, scene-conditioned router, action and future heads, retention probes, and metric-aware auxiliary branch. Provider outputs remain fixed conditioning features throughout WAM optimization.

The three expert-branch query banks contain 8 tokens each, and the scene vector $s_t$ is attention-pooled to width $d{=}1024$. The WAM is optimized with AdamW ($\beta{=}(0.9,0.95)$, weight decay $0.05$) at a peak learning rate of $1{\times}10^{-4}$ with 1k linear-warmup steps and cosine decay, at an effective batch size of 128, in bf16 precision; no augmentation is applied beyond the provider's preprocessing. Multi-seed results use training seeds $\{0,1,2\}$ (Appendix~\ref{app:diagnostics}). The flow-matching action objective samples the flow time $\tau\sim\mathcal{U}(0,1)$ and base noise $x_0\sim\mathcal{N}(0,\mathbf{I})$. Future targets are the frozen-encoder latents of the next four frames ($2$\,s at $2$\,Hz), normalized like $\mathbf F_t$; the action-free mode replaces the trajectory condition with a learned null token; and in the joint phase, $\mathbf a_{\mathrm{pred}}$ is the one-step endpoint estimate at the sampled $\tau$, the same construction as the branch drafts.

\paragraph{Inference}
The provider encodes the observation once; the router produces the scene-conditioned gates $\boldsymbol\alpha$ in a single feed-forward pass and combines the expert conditions, after which the action-free future head predicts $\hat{v}_{\mathrm{free}}$ once. The rectified-flow actor then integrates 25 Euler steps and returns one trajectory. The auxiliary metric branch, prior-retention probes, branch drafts, and action-conditioned future prediction are absent from inference. Thus, single-trajectory inference excludes candidate generation, scorer-based selection, reranking, and test-time search, but it still uses multiple numerical integration steps rather than a single network evaluation.

\section{Staged Optimization and Gradient Routing}
\label{app:training_details}

\paragraph{Two-phase WAM training}
The WAM is trained with the total objective in Eq.~\eqref{eq:total}. During the first $T_w$ steps, $\mathcal{L}_{\mathrm{act}}$ and $\mathcal{L}_{\mathrm{route}}$ are disabled and the future, auxiliary, and prior-retention objectives train their permitted paths; with the zero-initialized output layer of $g_r$, gating stays uniform throughout warmup. Joint training then activates the flow-matching action objective together with routing distillation, whose branch drafts depend on the action head's velocity predictions. The future-conditioning output projection is initialized to zero, so the newly attached future path initially behaves as an identity-preserving residual. Its influence is learned as joint optimization proceeds.

\paragraph{Objective formulations and branch drafts}
The future head supports an action-free and an action-conditioned mode,
\begin{equation}
\hat{v}_{\mathrm{free}}=\mathrm{FP}(\hat{c}_{\mathrm{vid}},\varnothing),
\qquad
\hat{v}(\mathbf{a})=\mathrm{FP}(\hat{c}_{\mathrm{vid}},\mathbf{a}),
\end{equation}
where $\varnothing$ denotes that no trajectory condition is supplied. Both predictions are $\ell_1$-regressed against the frozen-encoder target $v_{\mathrm{gt}}$, with direct action-conditioned supervision only at the demonstration $\mathbf{a}_{\mathrm{gt}}$:
\begin{equation}
\mathcal{L}_{\mathrm{fut}}=\tfrac{1}{2}\big(\lVert\hat{v}_{\mathrm{free}}-v_{\mathrm{gt}}\rVert_1+\lVert\hat{v}(\mathbf{a}_{\mathrm{gt}})-v_{\mathrm{gt}}\rVert_1\big)
\end{equation}

The auxiliary regressor predicts a $K$-dimensional quality vector from the scene vector, the flattened trajectory, and the pooled imagined outcome,
\begin{equation}
R(\mathbf{a})=g_\omega\big(s_t,\mathrm{flat}(\mathbf{a}),\mathrm{pool}(\hat{v}(\mathbf{a}))\big)\in[0,1]^{K}
\end{equation}
where $\mathrm{flat}(\cdot)$ vectorizes the trajectory and $\mathrm{pool}(\cdot)$ reduces the predicted future latent. It is trained on both in-batch predictions and offline-pool actions,
\begin{equation}
\begin{split}
\mathcal{L}_{\mathrm{aux}} &= \tfrac{1}{2}\,\mathrm{MSE}\big(R(\mathbf{a}_{\mathrm{pred}}),\hat{\mathbf{s}}\big)+\tfrac{1}{2}\,\mathrm{MSE}\big(R(\mathbf{a}_{\mathrm{pool}}),\mathbf{s}_{\mathrm{pool}}\big)
\end{split}
\end{equation}
whose score vectors $\hat{\mathbf{s}}$ and $\mathbf{s}_{\mathrm{pool}}$ are distinct from the scene vector $s_t$.

The actor is trained with the standard flow-matching objective: for a base-noise sample $x_0\in\mathbb R^{H\times3}$, a flow time $\tau\in[0,1]$, and the interpolation $x_\tau=(1-\tau)\,x_0+\tau\,\mathbf{a}_{\mathrm{gt}}$ with target velocity $\mathbf{a}_{\mathrm{gt}}-x_0$,
\begin{equation}
\mathcal{L}_{\mathrm{act}}=\big\lVert v_\theta\big(x_\tau,\tau,[\hat{c}_{\mathrm{act}};\hat{c}_{\mathrm{exp}}],\hat{v}_{\mathrm{free}}\big)-(\mathbf{a}_{\mathrm{gt}}-x_0)\big\rVert_2^2,
\end{equation}
and the actor receives the same $\hat v_{\mathrm{free}}$ construction during training and inference, avoiding a future-conditioning mismatch.

Branch drafts are one-step data-endpoint estimates in which branch $c$'s condition replaces the gated combination $\hat c_{\mathrm{exp}}$,
\begin{equation}\label{eq:draft}
\hat{\mathbf a}_c=\operatorname{sg}\!\big[x_\tau+(1-\tau)\,v_\theta\big(x_\tau,\tau,[\hat{c}_{\mathrm{act}};\hat{c}_{c}],\operatorname{sg}[\hat{v}_{\mathrm{free}}]\big)\big].
\end{equation}
Each draft is scored by the offline-pool interpolation, and $q_c$ is the mean of its $K$-dimensional score vector; the router is then supervised with the softened target gate $\alpha^{\ast}=\operatorname{softmax}(\mathbf q/T_r)$ via the cross-entropy
\begin{equation}
\mathcal{L}_{\mathrm{route}}=-\sum_{c}\alpha^{\ast}_{c}\log\alpha_{c}.
\end{equation}

\begin{table*}[!t]\centering
\caption{\textbf{Ablation on perception-prior acquisition} under NAVSIM v2 EPDMS evaluation on navtest. Selected sub-metrics are shown. Top: provider composition, retrained under the full protocol, including a single-teacher removal in which the provider is re-distilled without the dynamic teacher and the WAM is retrained. Bottom: inference-time removal of each expert prior (no retraining), probing the planner's reliance on it.}
\label{tab:abl_prior}
\setlength{\tabcolsep}{6pt}
\begin{tabular}{@{}l | rrrrrr >{\columncolor{pdmsgray}}r@{}}
\toprule
\textbf{Variant} & \textbf{NC$\uparrow$} & \textbf{DAC$\uparrow$} & \textbf{TTC$\uparrow$} & \textbf{EP$\uparrow$} & \textbf{LK$\uparrow$} & \textbf{EC$\uparrow$} & \textbf{EPDMS$\uparrow$} \\
\midrule
\rowcolor{oursblue}PerceptDrive (full) & \textbf{98.6} & \textbf{98.8} & \textbf{97.9} & \textbf{87.5} & \textbf{96.9} & \textbf{86.3} & \textbf{90.2} \\
w/o \texttt{[DYN]} distillation (retrained) & 98.4 & 98.6 & 97.7 & 85.9 & 96.7 & 84.2 & 88.6 \\
w/o prior distillation & 97.8 & 95.9 & 96.8 & 84.9 & 93.8 & 82.9 & 85.5 \\
w/o VLM stream & 96.9 & 94.6 & 95.7 & 83.0 & 92.4 & 81.0 & 81.9 \\
\midrule
w/o \texttt{[GEO]} & 98.3 & 96.4 & 97.6 & 87.1 & 94.7 & 85.9 & 88.1 \\
w/o \texttt{[SEM]} & 97.2 & 98.2 & 96.3 & 87.0 & 96.4 & 85.8 & 88.5 \\
w/o \texttt{[DYN]} & 98.2 & 98.4 & 97.4 & 85.2 & 96.5 & 83.6 & 88.3 \\
\bottomrule
\end{tabular}
\end{table*}

\paragraph{Offline supervision pool}
For each scene, the offline pool contains the demonstration trajectory, 16 smoothly perturbed copies, and a constant-velocity baseline, all scored by the privileged rule-based planner on navtrain. 
The planner assigns $K{=}8$ sub-scores to each trajectory---NC, DAC, DDC, TLC, EP, TTC, LK, and EC; HC is excluded because it depends on the executed history rather than the candidate. During warmup, $\mathcal{L}_{\mathrm{aux}}$ is evaluated on offline-pool actions. During joint training, it additionally receives in-batch predicted actions. Because those predictions need not coincide with a scored pool element, their supervision targets are interpolated from nearby scored trajectories using $k{=}5$ nearest neighbors.

Warmup covers the first $T_w{=}3{,}000$ steps; the joint objective uses $\lambda_f{=}0.2$, $\lambda_{\mathrm{ret}}{=}0.1$, and $\lambda_r{=}0.1$ with routing temperature $T_r{=}0.5$, and the retention weight remains fixed across both phases; branch drafts are supervised through the same $k{=}5$ nearest-neighbor interpolation as in-batch predictions.

$\hat v_{\mathrm{free}}$, $\hat v(\mathbf a)$, and $v_{\mathrm{gt}}$ share the same tensor shape, and the flow time $\tau$ is distinct from the planning index $t$. Action-conditioned predictions for perturbed or generated actions are extrapolations rather than simulator-equivalent counterfactuals; modeling multimodal futures is outside the present scope.

Perturbed pool copies are generated by adding zero-mean Gaussian offsets to spline control points (lateral $\sigma{=}0.5$\,m, longitudinal $\sigma{=}1.0$\,m, heading $\sigma{=}2^{\circ}$) followed by a cubic-spline refit, keeping the perturbations kinematically smooth. The $k$-nearest-neighbor interpolation uses Euclidean distance between flattened normalized waypoint vectors with inverse-distance weights; queries beyond the pool's 95th-percentile nearest-neighbor distance fall back to the single nearest element. Planned $2$\,Hz waypoints are upsampled to the evaluator frequency by cubic-spline interpolation. The fidelity of this interpolation and of the one-step drafts is quantified in Appendix~\ref{app:diagnostics}.

\paragraph{Gradient routes}
The five WAM objectives share a backbone but use deliberately separated gradient paths:
\begin{itemize}
  \item $\mathcal{L}_{\mathrm{act}}$ updates the action path and its shared representation. Its input $\hat{v}_{\mathrm{free}}$ is detached, so the action loss does not update the future head.
  \item $\mathcal{L}_{\mathrm{fut}}$ trains the action-free and action-conditioned future predictions against the same frozen-encoder target. Direct action-conditioned supervision is available at the demonstrated action; applying this predictor to perturbed or predicted actions is therefore an extrapolation.
  \item $\mathcal{L}_{\mathrm{aux}}$ receives detached action and future inputs. It updates the regressor and reaches the shared planning backbone only through the non-detached scene vector $s_t$; it does not reweight the imitation loss or optimize a trajectory directly.
  \item $\mathcal{L}_{\mathrm{ret}}$ uses detached pooled expert-slot representations as per-branch targets. Its gradients update the probes and propagate through the corresponding branch readouts $C_c$, shaping how each expert query bank reads the frozen provider without updating it.
  \item $\mathcal{L}_{\mathrm{route}}$ scores stop-gradient branch drafts with the privileged pool and distills the resulting target gates into $g_r$ via cross-entropy. It updates the router and reaches the shared backbone only through the non-detached $s_t$; the drafts, their scores, and $\alpha^{\ast}$ send no gradients into the action or future head.
\end{itemize}
These routes explain the design-choice ablation in Table~\ref{tab:abl_design}: detaching $s_t$ removes the only path by which metric supervision can shape the shared planning representation, and disabling $\mathcal{L}_{\mathrm{route}}$ leaves the gate with end-to-end signal only.

\section{Ablation Protocol and Interpretation}
\label{app:ablation_protocol}

The two blocks in Table~\ref{tab:abl_prior} answer different questions. The provider-composition variants are retrained under the full protocol and test whether the VLM stream and multi-teacher distillation contribute to the final system. Unless otherwise stated, all retrained variants use the same 30k-step WAM budget, random seed, and evaluation pipeline as the complete model, without variant-specific tuning. The expert-token variants remove one frozen provider token only at inference and do not retrain the WAM. They therefore probe sensitivity of the trained planner to each token, not an isolated causal effect of learning that prior; the per-metric visualization of this block in Figure~\ref{fig:ablation_viz}(b) is subject to the same interpretive scope.

Table~\ref{tab:abl_wam} uses nested WAM configurations: future conditioning is added to imitation, metric supervision and per-branch prior retention next, and metric-distilled routing last; variants without MR keep the three expert branches under uniform gating. Variants lacking an objective train only their remaining objectives during warmup, and the imitation-only variant proceeds directly to joint training since all warmup objectives are disabled. The monotonic trend supports the complete construction, but each change is conditional on the preceding configuration and should not be read as an order-independent component effect. Table~\ref{tab:abl_design} instead changes exactly one choice relative to the complete model. The first three rows test the zero-initialized future projection, staged warmup, and gradient path through $s_t$. The last three rows change the routing/fusion scheme and are retrained: the static variant replaces $g_r(s_t)$ with a learnable scene-agnostic $\operatorname{softmax}(w)$; the trajectory-averaging variant generates one trajectory per expert condition and averages them, an ablation-only protocol that requires three trajectories per scene and lies outside the single-trajectory inference used elsewhere; and the no-distillation variant keeps $g_r$ but disables $\mathcal{L}_{\mathrm{route}}$.

Two router-input controls isolate what the gate reads. The command-only router replaces $s_t$ with the command embedding alone, so its gate can express at most a per-command lookup; the no-command variant removes the command embedding from the router's input while keeping it available to the heads. The retention-design rows are likewise retrained: removing $\mathcal{L}_{\mathrm{ret}}$ from the full model (with routing kept) isolates the retention gain outside the nested order of Table~\ref{tab:abl_wam}; shuffled pairing permutes the branch--target assignment so each branch reconstructs a mismatched prior; the shared-target variant reconstructs the concatenated $\bar h$ from every branch, removing target specificity; and the token-level variant reconstructs per-token slot features instead of pooled means. The source-masked variant restricts each expert-branch bank's cross-attention to its designated slot group and the observation latents, replacing the default shared-pool access; its small drop suggests that hard partitioning forfeits useful shared context, while the retention objective already maintains specialization. Representation-level diagnostics for these variants appear in Appendix~\ref{app:diagnostics}.

\begin{table}[!t]\centering
\caption{Ablation on key design, routing, and retention choices. The first block changes one optimization choice, the second block changes the routing/fusion scheme or the router input, and the third block changes the retention design or the experts' read access; each row modifies exactly one item relative to the full model. PDMS uses NAVSIM v1 navtest; EPDMS uses NAVSIM v2 navtest.}
\label{tab:abl_design}
\small
\setlength{\tabcolsep}{3.5pt}
\begin{tabular}{l | r >{\columncolor{pdmsgray}}r}
\toprule
\textbf{Variant} & \textbf{PDMS$\uparrow$} & \textbf{EPDMS$\uparrow$} \\
\midrule
PerceptDrive (full) & \textbf{90.4} & \textbf{90.2} \\
w/o zero-initialized future projection & 86.3 & 86.4 \\
w/o warmup (single-phase) & 88.8 & 88.6 \\
w/o gradient through $s_t$ & 85.8 & 86.0 \\
\midrule
Static learnable weights (scene-agnostic) & 89.7 & 88.8 \\
Trajectory-level averaging fusion & 88.5 & 88.3 \\
w/o routing distillation ($\boldsymbol\alpha$ end-to-end) & 90.0 & 89.5 \\

Command-only router ($\boldsymbol\alpha=g_r(c_t)$) & 89.9 & 89.0 \\
w/o command in router input & 90.2 & 89.9 \\
\midrule

w/o $\mathcal{L}_{\mathrm{ret}}$ (retention removed) & 89.3 & 88.8 \\
Shuffled branch--target pairing & 89.1 & 88.6 \\
Shared concatenated target & 89.9 & 89.5 \\
Token-level retention targets & 90.3 & 90.1 \\
Source-masked expert reads & 90.0 & 89.7 \\
\bottomrule
\end{tabular}
\normalsize
\end{table}

\section{Results on the navhard Split}
\label{app:navhard}

\begin{table*}[!t]\centering
\caption{Comparison on the public NAVSIM v2 navhard benchmark under two-stage pseudo-simulation. Sub-scores are listed per stage and EPDMS is the final combined score; methods may differ in training setup and inference budget. \textbf{Bold} marks the best value in each column per stage.}
\label{tab:navhard}
\setlength{\tabcolsep}{4pt}
\begin{tabular}{l | c | rrrrrrrrr >{\columncolor{pdmsgray}}c}
\toprule
\textbf{Method} & \textbf{Stage} & \textbf{NC$\uparrow$} & \textbf{DAC$\uparrow$} & \textbf{DDC$\uparrow$} & \textbf{TLC$\uparrow$} & \textbf{EP$\uparrow$} & \textbf{TTC$\uparrow$} & \textbf{LK$\uparrow$} & \textbf{HC$\uparrow$} & \textbf{EC$\uparrow$} & \textbf{EPDMS$\uparrow$} \\
\midrule
\multirow{2}{*}{DiffusionDrive~\citep{diffusiondrive}} & 1 & 96.8 & 86.0 & 98.8 & 99.3 & 84.0 & 95.8 & \textbf{96.7} & 97.6 & \textbf{79.6} & \\
 & 2 & 80.1 & 72.8 & 84.4 & 98.4 & \textbf{85.9} & 76.6 & 46.4 & 96.3 & \textbf{72.8} & \multirow{-2}{*}{27.5} \\
\midrule
\multirow{2}{*}{MindDrive~\citep{minddrive}} & 1 & 96.1 & 86.0 & 98.8 & 99.3 & 83.3 & 95.6 & 94.4 & 97.6 & 74.7 & \\
 & 2 & \textbf{82.6} & \textbf{79.1} & \textbf{86.4} & 98.0 & 85.3 & \textbf{79.4} & \textbf{49.2} & 96.5 & 71.0 & \multirow{-2}{*}{30.5} \\
\midrule
\multirow{2}{*}{DriveLaW~\citep{drivelaw}} & 1 & 97.3 & 89.1 & \textbf{99.2} & \textbf{99.6} & \textbf{84.3} & \textbf{97.1} & 96.2 & \textbf{97.8} & 67.6 & \\
 & 2 & 82.5 & 67.6 & 83.5 & 98.1 & 84.8 & 78.5 & 45.8 & 96.4 & 57.3 & \multirow{-2}{*}{30.6} \\
\midrule
\rowcolor{oursblue} & 1 & \textbf{97.5} & \textbf{94.2} & 99.0 & 99.4 & 83.9 & 95.6 & 96.4 & 97.7 & 78.1 & \\
\rowcolor{oursblue}\multirow{-2}{*}{\textbf{PerceptDrive (Ours)}} & 2 & 81.9 & 71.4 & 85.1 & \textbf{98.5} & 84.5 & 77.9 & 48.7 & \textbf{96.6} & 69.3 & \multirow{-2}{*}{\textbf{34.5}} \\
\bottomrule
\end{tabular}
\end{table*}

We evaluate PerceptDrive on the public NAVSIM v2 navhard benchmark using two-stage pseudo-simulation, reporting per-stage sub-scores. PerceptDrive reaches 34.5 EPDMS, the best score in Table~\ref{tab:navhard}, 3.9 points above the strongest baseline, DriveLaW (30.6). Stage~1 shows strong collision avoidance and drivable-area compliance, while the stage-2 drops in LK and EC are shared by all listed methods and mark the remaining headroom of the benchmark.

\section{Statistical Stability and Extended Diagnostics}
\label{app:diagnostics}

\paragraph{Multi-seed stability}
Table~\ref{tab:multiseed} reports mean$\pm$std over three training seeds $\{0,1,2\}$ for the nested variants of Table~\ref{tab:abl_wam} and for the end-to-end-gate control, each evaluated with three base-noise seeds fixed before evaluation. The variant ordering is preserved by every seed, and the evaluation-seed standard deviation is at most $0.05$ for all variants, so the reported variability is dominated by training stochasticity. Per-scene paired bootstrap over navtest (10k resamples, seed-0 checkpoints) separates the full model from each ablated variant at the 95\% level, including the tightest comparison against the end-to-end gate.

\begin{table}[!t]\centering
\caption{Multi-seed stability: mean$\pm$std over three training seeds, three fixed base-noise seeds each. Rows follow Table~\ref{tab:abl_wam} plus the end-to-end-gate control of Table~\ref{tab:abl_design}.}
\label{tab:multiseed}
\small
\setlength{\tabcolsep}{3.5pt}
\begin{tabular}{l | c >{\columncolor{pdmsgray}}c}
\toprule
\textbf{Variant} & \textbf{PDMS$\uparrow$} & \textbf{EPDMS$\uparrow$} \\
\midrule
Imitation only & 86.9$_{\pm.17}$ & 84.6$_{\pm.26}$ \\
+ Future conditioning & 87.5$_{\pm.22}$ & 85.4$_{\pm.08}$ \\
+ Metric supervision & 88.6$_{\pm.06}$ & 87.1$_{\pm.15}$ \\
+ Prior retention & 89.8$_{\pm.15}$ & 88.7$_{\pm.17}$ \\
w/o routing distillation ($\boldsymbol\alpha$ end-to-end) & 90.0$_{\pm.21}$ & 89.5$_{\pm.16}$ \\
\rowcolor{oursblue}PerceptDrive (full) & \textbf{90.4}$_{\pm.05}$ & \textbf{90.2}$_{\pm.11}$ \\
\bottomrule
\end{tabular}
\normalsize
\end{table}

\paragraph{Mechanism generality across providers}
Table~\ref{tab:providers_mech} disentangles the planner-side mechanisms from provider strength by training the same WAM configurations on the QA-only provider, i.e., without multi-teacher distillation. Prior retention and metric-distilled routing improve both providers---$+1.8$ EPDMS on the QA-only provider and $+3.1$ on the distilled provider---indicating that the mechanisms are not an artifact of the provider's construction cost, while richer priors amplify their effect. The bottom-left cell coincides with the ``w/o prior distillation'' row of Table~\ref{tab:abl_prior}, which uses the full-mechanism WAM.

\begin{table}[!t]\centering
\caption{NAVSIM v2 EPDMS on navtest when the same planner-side mechanisms are trained on the QA-only provider (no multi-teacher distillation) versus the distilled provider. The distilled column reproduces Table~\ref{tab:abl_wam}; the QA-only full-mechanism cell reproduces Table~\ref{tab:abl_prior}.}
\label{tab:providers_mech}
\small
\setlength{\tabcolsep}{4.5pt}
\begin{tabular}{l | c >{\columncolor{pdmsgray}}c}
\toprule
\textbf{WAM configuration} & \textbf{QA-only} & \textbf{Distilled} \\
\midrule
FC + MS (uniform gating) & 83.7 & 87.1 \\
+ per-branch prior retention & 84.8 & 88.7 \\
\rowcolor{oursblue}+ metric-distilled routing (full) & \textbf{85.5} & \textbf{90.2} \\
\midrule
$\Delta$ from mechanisms & $+1.8$ & $+3.1$ \\
\bottomrule
\end{tabular}
\normalsize
\end{table}

\paragraph{Validity of the routing supervision}
Table~\ref{tab:proxy} quantifies the two approximations behind $\mathcal{L}_{\mathrm{route}}$. %
First, on held-out perturbations excluded from the offline pools (four per scene across 2{,}048 navtrain-dev scenes), the $k$-NN interpolated sub-scores track directly evaluated privileged scores with an overall Spearman $\rho$ of $0.91$; the collision-boundary terms NC and TTC are the hardest at $0.88$ and $0.87$. Second, on 500 dev scenes we compare the one-step branch drafts of Eq.~\eqref{eq:draft} against full 25-step per-branch rollouts scored identically: the top-scoring expert agrees in $86.2\%$ of scenes (Kendall $\tau=0.72$), and the resulting target gates differ by $0.06$ in mean $\ell_1$ distance. Target gates are stable to the draft construction (std $0.03$ across fixed $\tau$ values from $0.3$ to $0.9$ in steps of $0.1$, versus the sampled-$\tau$ default), and their mean entropy at convergence is $0.86$ nats (uniform $=\ln 3=1.0986$~nats): soft, but consistently non-uniform. These surrogates are used only to construct routing supervision on navtrain; all reported PDMS and EPDMS values are computed directly by the corresponding official NAVSIM evaluator on navtest or navhard.

\begin{table}[!t]\centering
\caption{Validity of the routing supervision on navtrain-dev. Top: $k$-NN interpolated sub-scores versus directly evaluated privileged scores on held-out perturbations. Bottom: one-step branch drafts versus full 25-step per-branch rollouts.}
\label{tab:proxy}
\small
\setlength{\tabcolsep}{3.0pt}
\begin{tabular}{l | rrrrrr | r}
\toprule
 & \textbf{NC} & \textbf{DAC} & \textbf{TTC} & \textbf{EP} & \textbf{LK} & \textbf{EC} & \textbf{All} \\
\midrule
Spearman $\rho$ & .88 & .90 & .87 & .95 & .92 & .89 & .91 \\
MAE & .052 & .048 & .055 & .028 & .036 & .045 & .041 \\
\bottomrule
\end{tabular}

\vspace{4pt}

\begin{tabular}{l | c}
\toprule
\textbf{Draft-versus-rollout statistic} & \textbf{Value} \\
\midrule
Top-1 expert agreement & 86.2\% \\
Kendall $\tau$ over expert rankings & 0.72 \\
Mean $\ell_1$ gap between target gates & 0.06 \\
Target-gate std across fixed $\tau$ grid & 0.03 \\
Mean target-gate entropy (uniform $=1.0986$) & 0.86 \\
\bottomrule
\end{tabular}
\normalsize
\end{table}

\paragraph{Retention diagnostics}
Table~\ref{tab:retention_diag} probes branch specialization directly on navtrain-dev. Without $\mathcal{L}_{\mathrm{ret}}$, the three branch readouts are largely redundant (mean pairwise CKA $0.76$; cross-reconstruction selectivity $0.08$); with it, cross-branch similarity drops to $0.52$ and each branch reconstructs its own prior far better than the others' ($0.83$ versus $0.41$ mean cosine, selectivity $0.42$). The attention pattern tells the same story: with $\mathcal{L}_{\mathrm{ret}}$, each expert bank concentrates $31\%$ of its cross-attention mass on its designated slot group, versus $9\%$ without it; a uniform read over the $1{,}560$-token pool ($1{,}048$ provider tokens and $512$ video latents) would place $8/1{,}560=0.51\%$ on the eight designated slots. Together with the shuffled-pairing and source-masked controls in Table~\ref{tab:abl_design}, this supports reading the retention loss as target-specific anchoring that induces specialized reads, rather than a generic auxiliary reconstruction load.

\begin{table}[!t]\centering
\caption{Retention diagnostics on navtrain-dev: mean pairwise CKA between branch readouts, held-out cross-reconstruction cosine for matched (diagonal) and mismatched (off-diagonal) branch--target pairs, their gap (selectivity), and the mean share of each expert bank's cross-attention mass on its designated slot group (a uniform read over the $1{,}560$-token pool would give $8/1{,}560=0.0051$).}
\label{tab:retention_diag}
\small
\setlength{\tabcolsep}{2.5pt}
\begin{tabular}{l | c c c c c}
\toprule
\textbf{Variant} & \textbf{CKA} & \textbf{Diag.} & \textbf{Off-diag.} & \textbf{Select.} & \textbf{Own attn.} \\
\midrule
with $\mathcal{L}_{\mathrm{ret}}$ & 0.52 & 0.83 & 0.41 & 0.42 & 0.31 \\
w/o $\mathcal{L}_{\mathrm{ret}}$ & 0.76 & 0.69 & 0.61 & 0.08 & 0.09 \\
\bottomrule
\end{tabular}
\normalsize
\end{table}

\paragraph{Routing dispersion within commands}
Beyond the command-grouped means of Figure~\ref{fig:ablation_viz}(a), the gates vary within commands: the mean within-command standard deviation of $\boldsymbol\alpha$ is $0.11$, and a command-only predictor explains $38\%$ of the gate variance on navtest, leaving the majority to scene features. This complements the command-only router control in Table~\ref{tab:abl_design}.

\paragraph{Hyperparameter sensitivity}
Table~\ref{tab:sensitivity} varies each routing and retention hyperparameter around its default. Performance is flat within the explored ranges (spread $\leq 0.4$ EPDMS), indicating that the defaults were not finely tuned to navtest.

\begin{table}[!t]\centering
\caption{Sensitivity of NAVSIM v2 EPDMS on navtest to the routing/retention hyperparameters; defaults in bold. PDMS trends match.}
\label{tab:sensitivity}
\small
\setlength{\tabcolsep}{4.5pt}
\begin{tabular}{l | c | c}
\toprule
\textbf{Hyperparameter} & \textbf{Values} & \textbf{EPDMS$\uparrow$} \\
\midrule
$\lambda_{\mathrm{ret}}$ & 0.05 / \textbf{0.1} / 0.2 & 89.9 / \textbf{90.2} / 90.0 \\
$\lambda_{r}$ & 0.05 / \textbf{0.1} / 0.2 & 89.9 / \textbf{90.2} / 90.0 \\
$T_r$ & 0.25 / \textbf{0.5} / 1.0 & 89.8 / \textbf{90.2} / 89.9 \\
Perturbations per scene & 8 / \textbf{16} / 32 & 89.8 / \textbf{90.2} / 90.2 \\
$k$ (nearest neighbors) & 3 / \textbf{5} / 9 & 90.0 / \textbf{90.2} / 90.0 \\
\bottomrule
\end{tabular}
\normalsize
\end{table}

\FloatBarrier

\begin{table}[!t]\centering
\caption{Training cost by stage (AMD MI308X-192G).}
\label{tab:cost}
\small
\setlength{\tabcolsep}{3.0pt}
\begin{tabular}{l | c c c}
\toprule
\textbf{Stage} & \textbf{Hardware} & \textbf{Time} & \textbf{GPU$\cdot$h} \\
\midrule
1a QA adaptation (full FT) & 8$\times$MI308X & 3.8\,h & 30.4 \\
1b Multi-teacher distillation & 8$\times$MI308X & 19.0\,h & 152.0 \\
2 WAM training (30k steps) & 2$\times$MI308X & 113.6\,h & 227.2 \\
\midrule
Total & --- & --- & 409.6 \\
\bottomrule
\end{tabular}
\normalsize
\end{table}

\section{Computational Cost and Parameters}
\label{app:cost}

\paragraph{Training cost}
Table~\ref{tab:cost} breaks down the training cost by stage. The complete system trains for approximately 409.6 MI308X GPU-hours: provider construction (Stages~1a/1b) accounts for 182.4, and WAM optimization for 227.2.

\paragraph{Parameters}
The frozen provider contributes 2.09B parameters and the frozen V-JEPA~2-L encoder 0.33B; the WAM adds 421M trainable parameters. The train-only modules (the metric-aware auxiliary branch and its pooling heads) are dropped at inference, leaving 2.82B active parameters.

\paragraph{Accuracy--latency trade-off}
Table~\ref{tab:euler} varies the number of Euler steps at inference without retraining. A 10-step solver loses only 0.1 points at 53\,ms, and even the single-step solver retains much of the system's benefit (88.2 PDMS, above the imitation-only and future-conditioned variants of Table~\ref{tab:abl_wam}), since the gate and the predicted future are solver-independent.

\begin{table}[!htbp]\centering
\small
\setlength{\tabcolsep}{4.5pt}
\begin{tabular}{c | c >{\columncolor{pdmsgray}}c c}
\toprule
\textbf{Euler steps} & \textbf{PDMS$\uparrow$} & \textbf{EPDMS$\uparrow$} & \textbf{Latency (ms)} \\
\midrule
1 & 88.2 & 87.5 & 44 \\
5 & 90.0 & 89.6 & 48 \\
10 & 90.3 & 90.1 & 53 \\
\textbf{25 (default)} & \textbf{90.4} & \textbf{90.2} & 68 \\
\bottomrule
\end{tabular}
\normalsize
\caption{Accuracy--latency trade-off over flow-solver steps (no retraining); latency measured end to end on one MI308X.}
\label{tab:euler}
\end{table}


\FloatBarrier
\section{Additional Visualization}
\label{app:visualizations}

\begin{figure*}[!tp]
  \centering
  \includegraphics[width=0.97\textwidth,height=0.90\textheight,keepaspectratio]{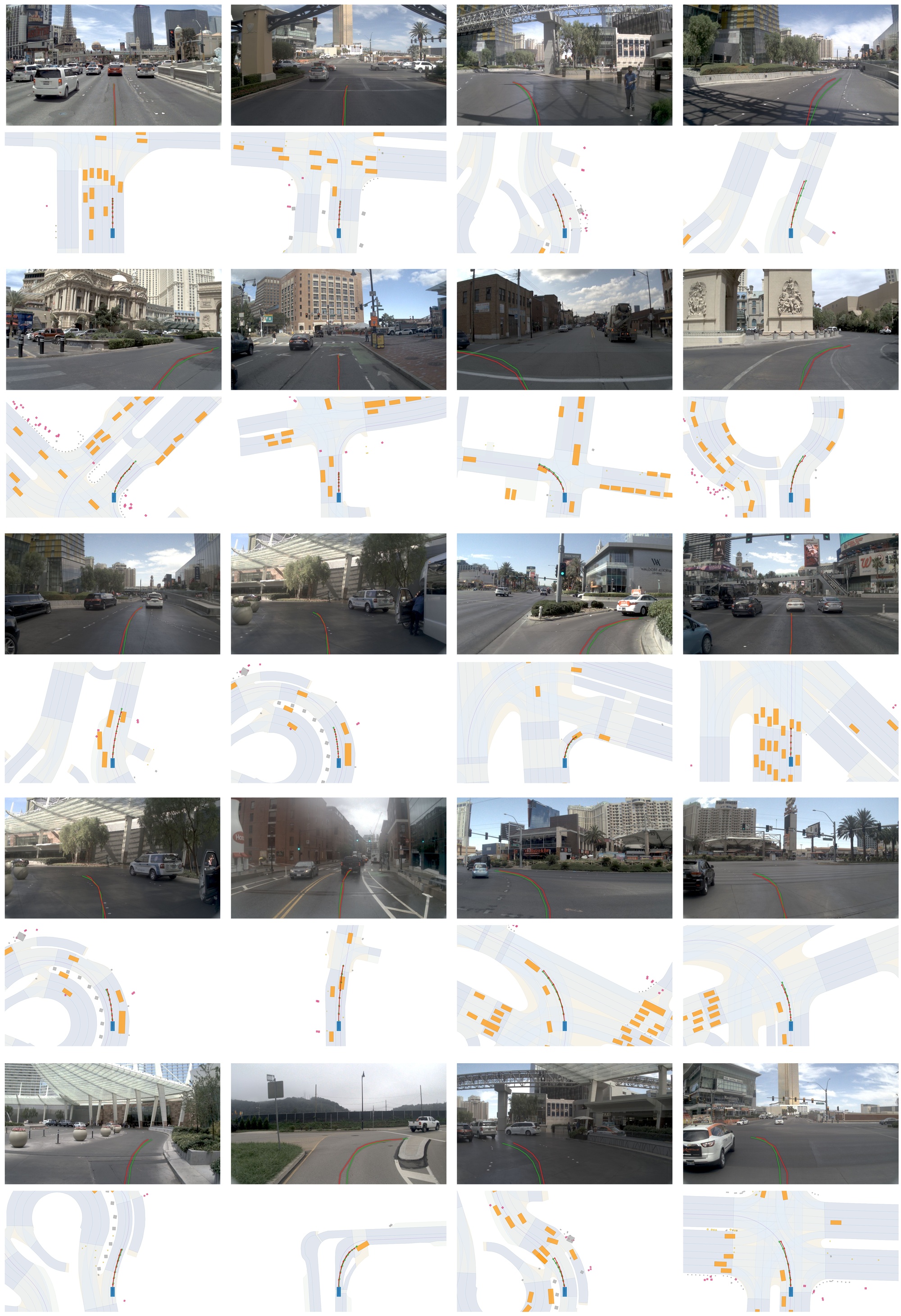}
  \caption{Additional qualitative results on NAVSIM navtest, in the layout of Figure~\ref{fig:qualitative_navsim}: each example pairs the front-view observation (top) with its BEV context and the generated trajectory (bottom). \trajectorylegend}
  \label{fig:app_qualitative_cases_1}
\end{figure*}

\begin{figure*}[!tp]
  \centering
  \includegraphics[width=0.97\textwidth,height=0.90\textheight,keepaspectratio]{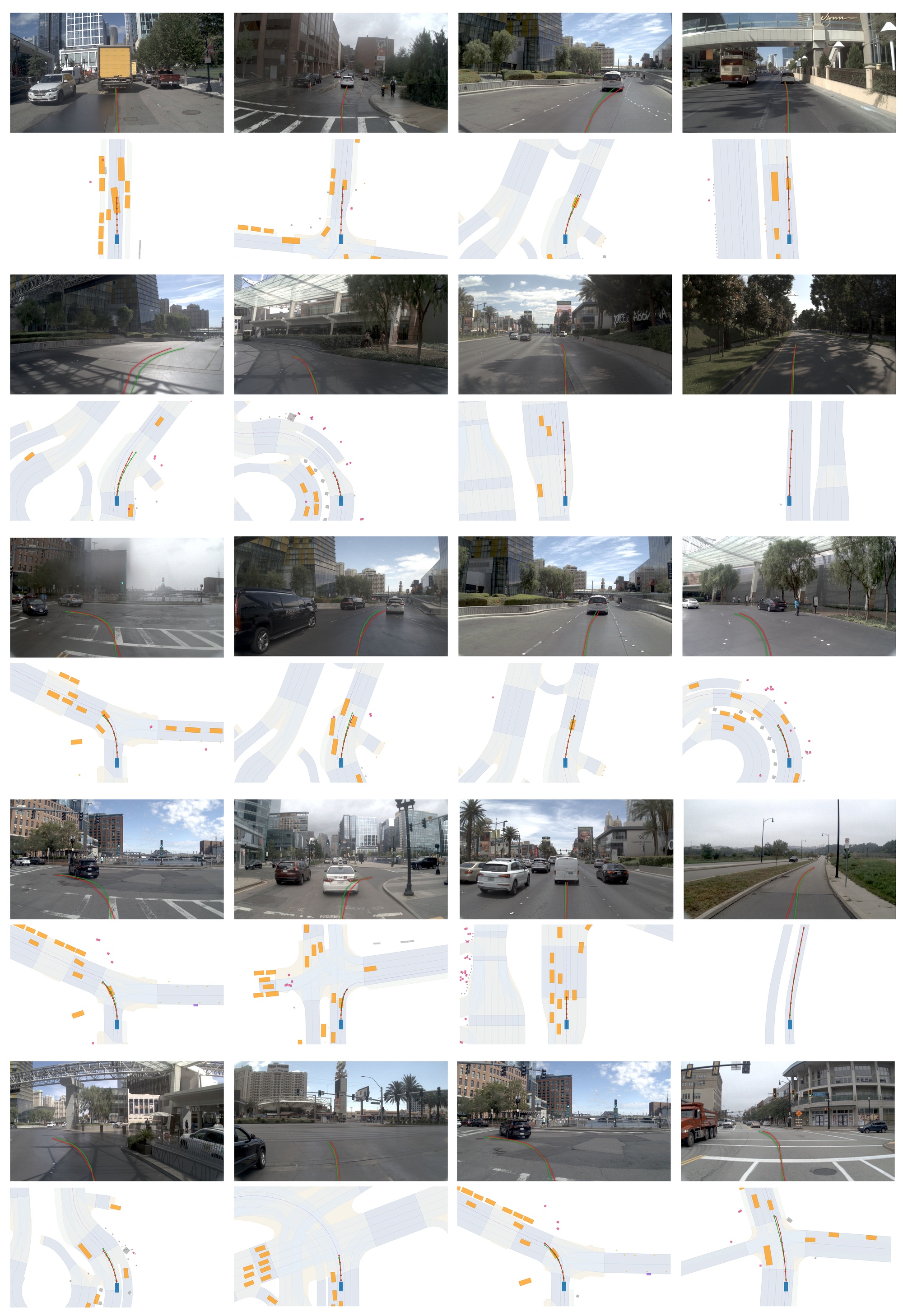}
  \caption{Additional qualitative results on NAVSIM navtest (continued): each example pairs the front-view observation (top) with its BEV context and the generated trajectory (bottom). \trajectorylegend}
  \label{fig:app_qualitative_cases_2}
\end{figure*}

Figures~\ref{fig:app_qualitative_cases_1} and~\ref{fig:app_qualitative_cases_2} show additional qualitative results across intersections, turns, lane changes, and dense traffic. The generated trajectories remain close to the logged paths across these scenarios, complementing the quantitative comparisons in the main paper.

\end{document}